%% file: dual_prompt_camera_ready.tex
\documentclass[runningheads]{llncs}
\usepackage{graphicx}
\usepackage{tikz}
\usepackage{comment}
\usepackage{amsmath,amssymb} % define this before the line numbering.
\usepackage{color}

\usepackage[accsupp]{axessibility}  % Improves PDF readability for those with disabilities.

\usepackage{microtype}
\usepackage{subfigure}
\usepackage{booktabs} % for professional tables
\usepackage{soul}

\usepackage[ruled,vlined]{algorithm2e}

\usepackage{xspace}
\usepackage{multirow}
\usepackage{caption}
\usepackage{sidecap}
\usepackage{array}
\usepackage[pagebackref,breaklinks,colorlinks,bookmarks=false]{hyperref}
\usepackage[capitalize,noabbrev]{cleveref}

\input{math_commands.tex}

\newcommand{\method}{DualPrompt\xspace}
\newcommand{\gprompt}{G-Prompt\xspace}
\newcommand{\eprompt}{E-Prompt\xspace}

\newcommand{\ie}{\emph{i.e.}\xspace}
\newcommand{\eg}{\emph{e.g.}\xspace}

\def\pfunc{{f_{\text{prompt}}}}
\def \iep{\texttt{start}_{\ve}}
\def \jep{\texttt{end}_{\ve}}
\def \igp{\texttt{start}_{\vg}}
\def \jgp{\texttt{end}_{\vg}}

\begin{document}
% \linenumbers
\pagestyle{headings}
\mainmatter
\def\ECCVSubNumber{5428}  % Insert your submission number here

\title{DualPrompt: Complementary Prompting for Rehearsal-free Continual Learning} % Replace with your title

\begin{comment}
\titlerunning{ECCV-22 submission ID \ECCVSubNumber} 
\authorrunning{ECCV-22 submission ID \ECCVSubNumber} 
\author{Anonymous ECCV submission}
\institute{Paper ID \ECCVSubNumber}
\end{comment}
%******************

%\begin{comment}
\titlerunning{Complementary Prompting for Rehearsal-free Continual Learning}
\author{Zifeng Wang\textsuperscript{1}\thanks{Work done while the author was an intern at Google Cloud AI Research. Email:~\email{zifengwang@ece.neu.edu}}, Zizhao Zhang\textsuperscript{2}, Sayna Ebrahimi\textsuperscript{2}, Ruoxi Sun\textsuperscript{2}, \\ Han Zhang\textsuperscript{3}, Chen-Yu Lee\textsuperscript{2}, Xiaoqi Ren\textsuperscript{2}, Guolong Su\textsuperscript{3}, \\ Vincent Perot\textsuperscript{3}, Jennifer Dy\textsuperscript{1}, and Tomas Pfister\textsuperscript{2}}
\authorrunning{Z. Wang, Z. Zhang et al.}
\institute{\textsuperscript{1}Northeastern University \quad \textsuperscript{2}Google Cloud AI \quad  \textsuperscript{3}Google Research}
%\end{comment}
%******************
\maketitle
\begin{abstract}
Continual learning aims to enable a single model to learn a sequence of tasks without catastrophic forgetting. 
Top-performing methods usually require a rehearsal buffer to store past pristine examples for experience replay, which, however, limits their practical value due to privacy and memory constraints.
In this work, we present a simple yet effective framework, \method, which learns a tiny set of parameters, called \emph{prompts}, to properly instruct a pre-trained model to learn tasks arriving sequentially without buffering past examples. 
\method presents a novel approach to attach complementary prompts to the pre-trained backbone, and then formulates the objective as learning task-invariant and task-specific ``instructions". 
With extensive experimental validation, \method consistently sets state-of-the-art performance under the challenging class-incremental setting. 
In particular, \method outperforms recent advanced continual learning methods with relatively large buffer sizes. 
We also introduce a more challenging benchmark, Split ImageNet-R, to help generalize \emph{rehearsal-free} continual learning research. Source code is available at \url{https://github.com/google-research/l2p}.
\end{abstract}

\section{Introduction}
The central goal of continual learning (CL) is to learn a sequence of tasks with a single model without suffering from \emph{catastrophic forgetting} \cite{mccloskey1989catastrophic} -- a significant  deterioration in performance on previously seen data. Many existing methods aim at preserving and extending the acquired knowledge during the continual learning process~\cite{hadsell2020embracing,mai2021online}. Architecture-based methods assign isolated parameters to encode learned knowledge from different tasks~\cite{li2019learn,loo2020generalized,mallya2018packnet,serra2018overcoming,wang2020learn}. However, they often introduce a substantial number of additional parameters and sometimes involve simplified assumption like known test time task identity~\cite{ebrahimi2020adversarial,mallya2018packnet,mallya2018piggyback}, which falls into the setting of task-incremental learning. 
However, the task-incremental setting is usually considered over-simplified~\cite{buzzega2020dark,mai2021online,masana2020class}, since task identity is not known at test time in the real world. 
Our work focuses on more difficult class-incremental setting with unknown test-time task identity. 
Another line of work, rehearsal-based CL methods, preserve past knowledge directly by keeping data from prior tasks in a rehearsal buffer~\cite{buzzega2020dark,cha2021co2l,pham2021dualnet}. 
Due to their conceptual simplicity, generalizability to various settings, and superior ability to mitigate catastrophic forgetting, rehearsal-based methods have been widely recognized as the reigning state-of-the-art~\cite{chaudhry2020using,buzzega2020dark} in the challenging class-incremental setting. 
Nevertheless, the dependence on rehearsal buffer has been criticised in the community \cite{serra2018overcoming,hadsell2020embracing,prabhu2020gdumb,lomonaco2020rehearsal}. 
While the performance of these methods is sensitive to the size of the buffer, GDumb~\cite{prabhu2020gdumb} argues that performing supervised training directly on a relatively large buffer already surpasses most recent CL methods. 
Critically, these methods cannot be used in applications with privacy concerns \cite{shokri2015privacy} or when memory budget is highly constrained \cite{smith2021memory}. 
Thus, it is desirable to develop a parsimonious, rehearsal-free continual learning method that can achieve similar or higher level of performance.

\begin{figure}[t]
\centering 
\includegraphics[width=0.85\linewidth]{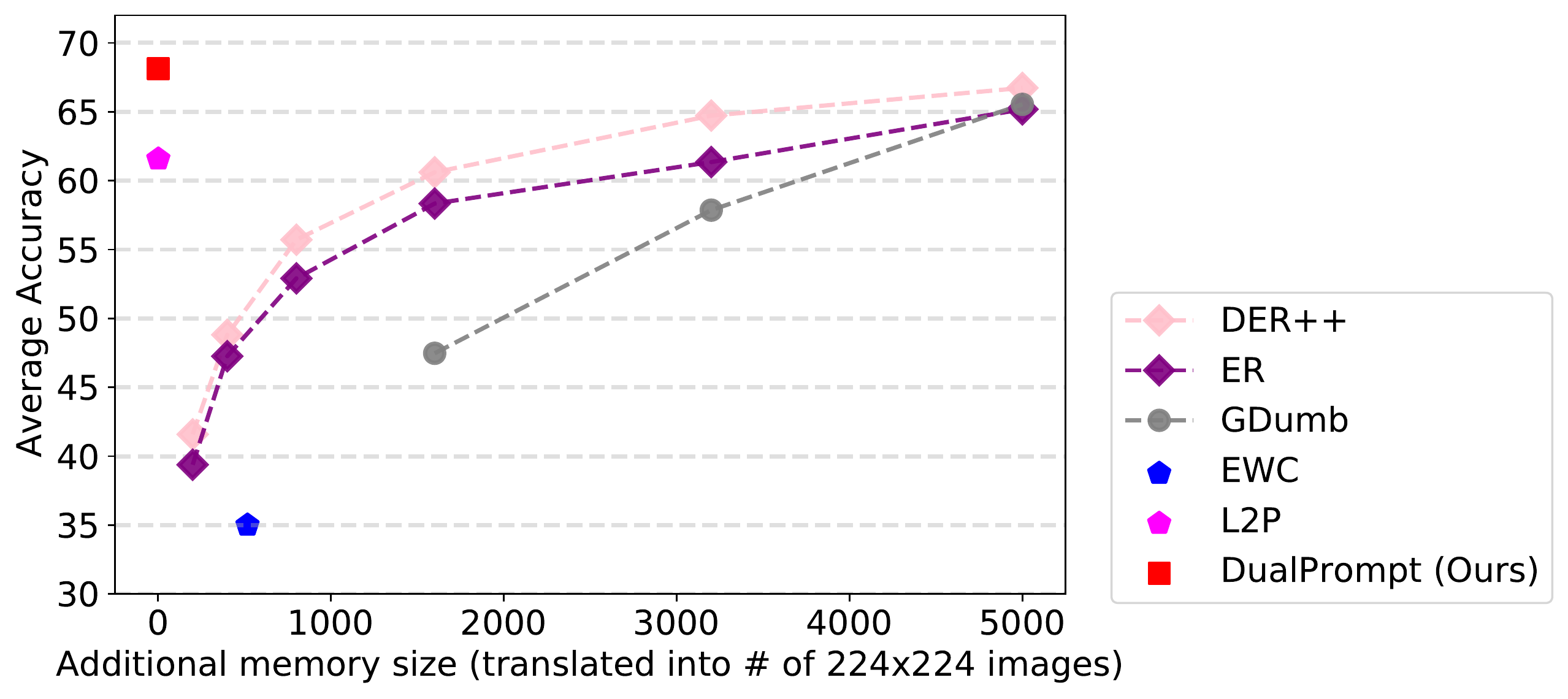}
%\vspace{-.3cm}
\caption{\smaller The average accuracy comparison on Split ImageNet-R, suggesting that the accuracy degradation is significant for representative rehearsal-based methods like DER++~\cite{buzzega2020dark} and ER~\cite{chaudhry2019tiny} when buffer size shrinks. 
Notably, they require a large rehearsal buffer (5000 images $\approx 20\%$ of the whole training set) to close the gap to our method.
In contrast, GDumb~\cite{prabhu2020gdumb} matches their result by training only on the i.i.d sampled buffer, without continual learning.
The additional parameters size required by \method is only about the bytes of one $224{\times}224$ RGB image. See Experiments section for discussion on compared methods.}
\label{fig:intro} %\vspace{-.8cm}
\end{figure}

A recent method, Learning to Prompt (L2P) \cite{wang2021learning} approaches this problem from a brand-new perspective -- it proposes to leverage learnable \emph{prompt} parameters to encode knowledge in a much more succinct way (i.e. prompt pool) than buffer, thus a rehearsal buffer is no longer necessary. Prompt techniques are originally introduced in natural language processing (NLP) for task adaptation \cite{liu2021pre} of large-scale pre-trained models by attaching fixed or learnable ``instructions'', since prompts are designed to instruct the model to properly reuse learned representations, instead of learning new representations from scratch. L2P successfully formulates the problem of learning new tasks as training small prompt parameters attached to a pre-trained frozen model. %Catastrophic forgetting is greatly mitigated.
%Although L2P takes an exciting step towards rehearsal-free continual learning, L2P tries to encode all learned knowledge in a shared memory of prompts, which inevitably leads to knowledge interference, and limited learning capacity.
L2P takes an exciting step towards rehearsal-free continual learning, although the performance is still lower than rehearsal-based methods. 
% The prompt pool in L2P is expected to make task-invariant and task-specific knowledge being learned jointly, implicitly encoded in this shared memory. \sayna{I don't think it was "expected" back then because you were not thinking of it as two things, isn't it better to say it as how you think of it now? how about this: 

In L2P, one single prompt pool is designed to transfer knowledge from one task to another without distinguishing between the common features among all tasks versus the features that are unique to each task. We argue such a design could be sub-optimal from the perspective of theory of Complementary Learning Systems (CLS)~\cite{mcclelland1995there,kumaran2016learning}, an intuition that many recent advanced CL methods are based on \cite{chaudhry2019tiny,pham2021dualnet,buzzega2020dark,arani2021learning}.
CLS suggests that humans learn continually via the synergy between two learning systems: the hippocampus focuses on learning pattern-separated representation on specific experiences, and the neocortex focuses on learning more general and transferable representation from past experience sequences. Thus, they are able to learn task-specific knowledge separately without interference while leveraging task-invariant knowledge to have greater learning capacity to learn future tasks better. %There are large body of recent works follow the CLS principle to design specific mechanism to explicitly manage the two kinds of experiences. 
However, previous CLS-driven methods still decouple or expand the backbone parameters learn the two kinds of knowledge~\cite{ebrahimi2020adversarial,pham2021dualnet,pham2020contextual}.
Thus, they still rely on constructing a rehearsal buffer repeatedly to consolidate decoupled knowledge to prevent catastrophic forgetting. 
%From a biological perspective, it is unrealistic to maintain any number of pristine observations or samples \cite{hadsell2020embracing}. 
%\zizhaoz{This paragraph did not clarify why CLS based CL methods can do better and why they still need to reply on a rehearsal buffer. Is that because CLS is not a solution for mitigating forgetting?}\sayna{I agree with Zizhao plus that I think this is s a bold claim. is it 100\% accurate? cause I think each one us have quite a lot of vivid memories - no matter how long ago - that we have stored in our memory and we do remember quite well. so it is realistic but maybe what's not realistic is that we can only maintain a limited number? - it's just that I think it is safe to not make any specific biology related comment because we are not expert in that and the paper is also not about that }

In this paper, we present \method, a rehearsal-free continual learning approach to explicitly learn two sets of disjoint prompt spaces, \emph{G(eneral)-Prompt} and \emph{E(xpert)-Prompt}, that encode task-invariant and task-specific instructions, respectively. \method directly decouples the higher-level prompt space, which turns out to be more effective and memory efficient that conventional methods which focus on the lower-level latent representation space.
%\method consists of two types of prompts: The \emph{G(eneral)-Prompt} that is shared among all tasks to learn task-invariant knowledge, and the \emph{E(xpert)-Prompt}, a set of prompts that are separately maintained for different tasks to learn task-specific knowledge.
% Our method further presents a highly modularized, unified framework to attach both type of prompts, which offers flexible ways to steer a pre-trained backbone model to learn new tasks without forgetting the old ones, and to encourage effective knowledge sharing between tasks. 
We further explore where and how to attach both types of prompts is crucial to steer the backbone model to learn with less forgetting and to achieve effective knowledge sharing, thus significantly enhancing the effectiveness of continual learning. 
% opens to future advances in prompting research for improving prompt-based continual learning.
% modify this paragraph to give more insights
 
% \zifeng{Different from conventional architecture- or rehearsal-based methods that allocates substantially large amount of memory for learning additional feature representation or saving raw data, \method continually learns high-level instructions to \emph{instruct} the pre-trained model to learn sequentially. Given that pre-trained models are widely applied common assets in the the machine learning community, it is also of great importance that \method is able to leverage the power of pre-trained models \emph{non-trivially} in CL. Moreover, unlike prior CLS-based methods which focus on decoupling the lower-level latent feature space, \method directly decouple the higher-level prompt space, which turns out to be more effective and memory efficient. In contrast to L2P, which simply attach all prompt parameters collectively to the input embedding layer, \method, as a unified framework, sheds light on how to attach prompts to the pre-trained backbone in a principled way. For example, our framework explores where and how to attach both type of prompts significantly influence the effectiveness of continual learning and opens to future advances in prompting research for improving prompt-based continual learning.}

Moreover, we introduce Split ImageNet-R, a new CL benchmark based on ImageNet-R \cite{hendrycks2020many} to the community. The intra-class diversity for each task in Split ImageNet-R is large (see Appendix~\ref{app:split_imr} for representative examples), thus a small buffer is not sufficient to represent past experiences. Figure~\ref{fig:intro} showcases that the size of rehearsal buffer needed is non-trivial for even advanced methods to perform well. 
% Figure~\ref{fig:intro} illustrates the performance of a number of recent CL method for a range of rehearsal buffer sizes. 
While rehearsal-based methods require a large buffer (up to $20\%$ of total training data) to achieve a competitive average accuracy, our method \method shows superior performance despite not using any rehearsal buffer.

In summary, our work makes the following contributions:
\begin{itemize}
    \item We propose \method, a simple and effective rehearsal-free CL method, comprised of \gprompt and \eprompt for learning task-invariant and task-specific knowledge, respectively. The method is fairly simple to apply without data or memory access concerns which is favorable for real-world CL scenarios.
    
    % \item \method presents a highly modularized prompt attaching framework to incorporate two types of prompts into the pre-trained models. For the first time, we empirically discover that properly attaching prompts to the backbone model is crucial to the effectiveness of continual learning.
    \item \method explores various design choices to incorporate these two types of prompts into the pre-trained models. For the first time, we empirically discover that properly attaching prompts to the backbone model is crucial to the effectiveness of continual learning.

    \item  We introduce a new CL benchmark, Split ImageNet-R to help validate the method. %\zifeng{Split ImageNet-R has large intra-class diversity, is closer to real world scenarios, and serves as a good alternative to ImageNet for ImageNet pre-trained models.} 
    \method sets new state-of-the-art performance on multiple benchmarks under the challenging class-incremental setting, and beats rehearsal-based methods with relatively large buffer size. 
\end{itemize}

\section{Related work}
%\vspace{-.2cm}
\textbf{Continual learning.} We discuss three related categories of continual learning methods: regularization-based, rehearsal-based and architecture-based methods.

\textit{Regularization-based methods} \cite{kirkpatrick2017overcoming,zenke2017continual,li2017learning,aljundi2018memory} address catastrophic forgetting by regularizing important parameters for learned tasks. Although these methods mitigate forgetting under simpler task-incremental setting, their performance under more challenging class-incremental setting \cite{mai2021online}, or more challenging datasets \cite{wu2019large} is not satisfactory.

\textit{Architecture-based methods} assign isolated parameters for each task. These methods can be further categorized as expanding the model \cite{rusu2016progressive,yoon2017lifelong,li2019learn,loo2020generalized,zhao2022deep}, or dividing the model \cite{mallya2018packnet,serra2018overcoming,wang2020learn,ke2020continual,ebrahimi2020adversarial}. However, a major part of the work is limited to the task-incremental setting \cite{serra2018overcoming,mallya2018packnet,mallya2018piggyback,ke2020continual}, while other work only considers specific convolutional-based architectures \cite{wortsman2020supermasks,pham2020contextual,ebrahimi2020adversarial}. However, \method aims at more challenging class-incremental setting, and focus on pre-trained transformer-based models. Moreover, architecture-based method generally require substantially large amount of additional parameters to assist model separation \cite{wang2020learn,ke2020continual,yan2021dynamically}. On the contrary, \method is much lightweight and only require negligible amount of parameters ($0.2\% - 0.6\%$ of full model size).

\textit{Rehearsal-based methods} save data from learned tasks in a rehearsal buffer to train with the current task. Although these methods share this quite simple idea, they are very effective even in the class-incremental setting. Several advanced rehearsal-base methods achieve state-of-the-art performance \cite{buzzega2020dark,cha2021co2l}. However, rehearsal-based methods deteriorates when buffer size \cite{cha2021co2l} decreases, and are eventually not applicable to data privacy sensitive scenarios \cite{shokri2015privacy}. 
Some recent methods are inspired from the Complementary Learning Systems (CLS). However, ACL~\cite{ebrahimi2020adversarial} is limited to the task-incremental setting, DualNet~\cite{pham2021dualnet} requires specific architecture design, and both methods still rely on a rehearsal buffer to work well.
Our \method tackles continual learning from a rehearsal-free perspective, standing upon a wise utilization of pre-trained models, thus getting rid of the shortcomings of rehearsal-based methods.

\noindent
\textbf{Prompt-based learning.}
As an emerging transfer learning technique in natural language processing (NLP), prompt-based learning (or prompting), applies a fixed function to condition the model, so that the language model gets additional instructions to perform the downstream task. However, the design of a prompting function is challenging and requires heuristics. To this end, recent work propose to apply prompts as learnable parameters, achieving outstanding performance on transfer learning \cite{lester2021power,li2021prefix}. Prompts capture task-specific knowledge with much smaller additional parameters, than its competitors, such as Adapter~\cite{wang2020k,pfeiffer2020adapterfusion} and LoRA~\cite{hu2021lora}. As discussed above, L2P \cite{wang2021learning} is the only work that connects prompting and continual learning. Differently, \method takes inspiration from CLS and presents a different approach to attach complementary prompts to the pre-trained backbone to learn task-invariant and task-specific instructions. We show \method outperforms L2P consistently.

% \zizhaoz{Does this a bit repeating intro?}
% \zifeng{Only keep result discussion.}
% A recent work, L2P \cite{wang2021learning} brings the idea of prompting into continual learning by learning a single set of prompts and organize them in key-value based pairs. Although L2P achieves great performance even compared to rehearsal-based methods, it still requires the rehearsal buffer to actually outperform them. \method takes inspiration from CLS and decouple the prompt parameters to learn task-invariant and task-specific knowledge separated, and further organizes them into a flexible framework, \combiner. As a result, \method not only outperforms L2P, but also beats rehearsal-based methods even with an unreasonable large buffer (training with only samples in buffer leads to good performance!), \emph{without storing any past examples}.
%\vspace{-.3cm}
\section{Prerequisites}
%\vspace{-.1cm}
\subsection{Continual learning problem setting}
Continual learning is defined as training machine learning models on a continuum of data from a sequence of tasks. We denote the sequence of tasks as $\mathcal{D} = \{\mathcal{D}_1, \cdots, \mathcal{D}_T\}$, where the $t$-th task $\mathcal{D}_t=\{(\vx_{i, t}, y_{i, t})\}_{i=1}^{n_t}$ contains tuples of the input sample $\vx_{i, t} \in \mathcal{X}$ and its corresponding label $y_{i, t} \in \mathcal{Y}$. 
The model $f_\theta: \mathcal{X} \to \mathcal{Y}$ is parameterized by $\theta$, such that it predicts the label $y = f_\theta(\vx) \in \mathcal{Y}$ given an unseen test sample $\vx$ from arbitrary tasks. 
Data from the previous tasks is not available when training future tasks. 

We use the widely-adopted assumption that the task boundaries are clear and the task switch is sudden at training time \cite{chaudhry2018efficient,pham2021dualnet}. Moreover, we consider the more challenging class-incremental learning~\cite{chaudhry2018efficient} setting, \ie, task identity is unknown for each example at test time. Also, following the settings in prior work \cite{wang2021learning}, we assume a pre-trained sequence model, \eg, a vision transformer (ViT) \cite{vit} on ImageNet, is available, a wide-used assumption in recent literature of the computer vision community. 
Unlike many rehearsal-based methods \cite{chaudhry2019tiny,buzzega2020dark}, we do not assume any form of rehearsal buffer as a prerequisite.

% a section for review prompt-based learning techniques.
%\vspace{-.3cm}
\subsection{Prompt-based learning} \label{sec:prompt}
%\vspace{-.1cm}
Prompt-based learning (or prompting) was first proposed in NLP for transfer learning.
The main idea of prompting is to add extra instruction for pre-trained models to perform downstream tasks conditionally \cite{liu2021pre}. Prompt Tuning~\cite{lester2021power}, one of the recent emerging techniques, proposes to attach a set of prompt parameters to frozen transformer-based language models \cite{raffel2020exploring} to perform downstream NLP tasks. The prompts are usually prepended to the input sequence to instruct the model prediction. We briefly illustrate the idea of Prompt Tuning below.

%\zizhaoz{Check if the content is too similar to L2P paper, and if some info that was used to assit L2P is still useful here. }
As we mainly focus on vision-related continual learning setting, here we introduce the definition of Prompt Tuning using the vision transformer (ViT) based sequence models~\cite{dosovitskiy2020image,vaswani2017attention}. 
In ViT, the input embedding layer transforms the input image into a sequence-like output feature $\vh \in \sR ^{L \times D}$, where $L$ is the sequence length and $D$ is the embedding dimension. When solving downstream tasks, the pre-trained backbone is kept frozen as a general feature extractor, and the prompt parameters $\vp \in \sR ^{L_{\vp} \times D}$ with sequence length $L_{\vp}$ and embedding dimension $D$ are prepended to the embedding feature along the sequence length dimension to form the extended embedding feature. Finally, the extended feature is sent to the rest of the model for performing classification tasks. Prompt serves as a lightweight module to encode high-level instruction to instruct the backbone to leverage pre-trained representations for downstream tasks.

%\sayna{it is not super clear why we always refer to as what prompt does as "instruction" - it's a bit weird to call them that way because when we talk about other parameters of the network we imply that they are "learners" so how did the prompts become instructors?}

% Learning to prompt (L2P) \cite{wang2021learning} first introduces prompting to the field of continual learning. L2P extends a single prompt parameter to a key-value paired set of prompts. 
\begin{figure}[t]
\centering 
\includegraphics[width=0.95\linewidth]{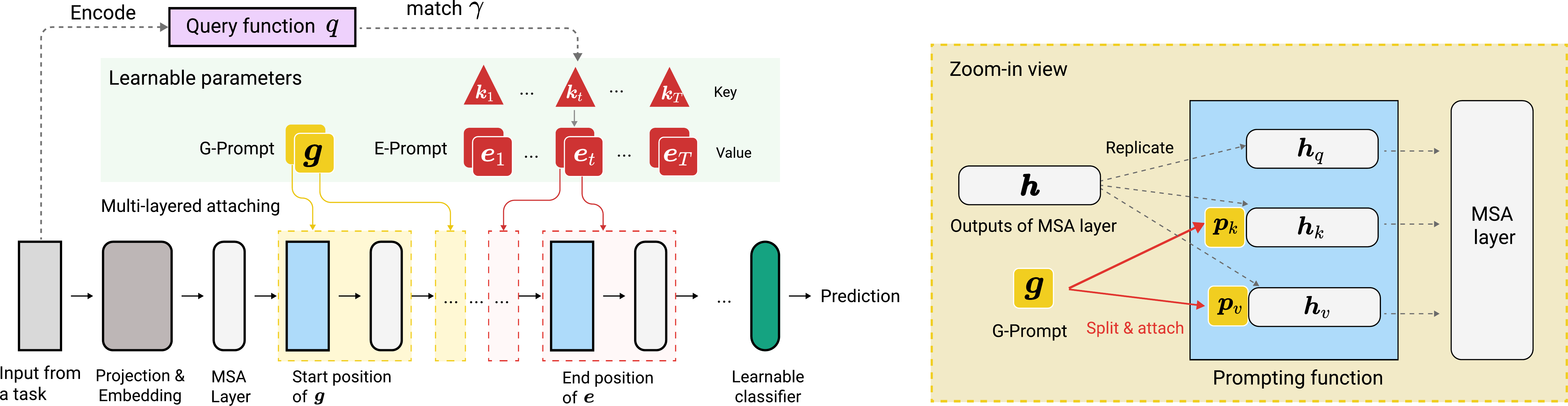}
% https://www.figma.com/file/y87HDSL0XmtTfDwlAppTRO/DualPrompts?node-id=2%3A3
\caption{\smaller Overview of \method. \textbf{Left}: At test time, an input is transformed by a query function to match the closest task key $\vk_t$ and the corresponding \eprompt $\ve_t$. Then the shared G(eneral)-Prompt $\vg$ and the matched E(xpert)-Prompt $\ve_t$ are attached to multiple MSA layers of a pre-trained transformer. 
At training time, the \eprompt is selected by task identity and the selected \eprompt and \gprompt are trained together with the classifier. 
\textbf{Right}: A prompting function is illustrated where the G-prompt is split equally and attached to the key and value replicas of the hidden feature (see Section~\ref{sec:combiner}) before passing them to the preceding MSA layer.
%The left shows the shared \gprompt, key-value paired \eprompt, and how to attach them to a single pre-trained self-attention layer. The right shows the architecture we use in our experiments that generated by our proposed \combiner framework (Section~\ref{sec:combiner}), where \gprompt and \eprompt are extended to attach to different multiple layers, with the Pre-T prompting function.
} 
%\vspace{-.3cm}
\label{fig:overview}
\end{figure} 

%\vspace{-.2cm}
\section{\method}
%\vspace{-.2cm}
Our proposed method, \method is illustrated in Figure~\ref{fig:overview}. We first introduce the complementary learning components, G- and E-prompts, in Section~\ref{sec:g-e-prompt} by showcasing how they work with a single multi-head self-attention (MSA) layer. 
% We then illustrate how to systematically attach two types of prompts via the modularized prompt attaching framework (\combiner) in Section~\ref{sec:combiner}.
We then explore design choices of attaching prompts to the backbone Section~\ref{sec:combiner}.
We finally present the overall objective for \method in Section~\ref{sec:main_obj}.

%\vspace{-.2cm}
\subsection{Complementary G-Prompt and E-Prompt} \label{sec:g-e-prompt}
%\vspace{-.1cm}
Given a pre-trained ViT $f$ with $N$ consecutive MSA layers, we further extend the notations introduced in \ref{sec:prompt} by denoting the input embedding feature of the $i$-th MSA layer as $\vh^{(i)}, i = 1, 2, \cdots, N$.

% \zizhaoz{Too many summarizing sentences! You already did in the start of Section 4.} We then demonstrate how to add \gprompt or \eprompt to input embedding features of an arbitrary layer to capture task-invariant and task-specific knowledge in the continual learning process. We will illustrate our method from a high-level perspective and leave the design details in the following subsection. 

\textbf{\gprompt:} $\vg \in \mathbb{R}^{L_g \times D}$ with sequence length $L_g$ and embedding dimension $D$, is a shared parameter for all tasks. Suppose we would like to attach \gprompt to the $i$-th MSA layer, 
\gprompt transforms $\vh^{(i)}$ via a \emph{prompting function}:
%\vspace{-.5mm}
\begin{equation} \label{eq:gprompt}
    \vh_g^{(i)} = \pfunc \left(\vg, \vh^{(i)}\right),
    %\vspace{-.7mm}
\end{equation}
where $\pfunc$ defines the approach how to attach the prompts to the hidden embeddings. %, \eg, the prepending operation mentioned in Section~\ref{sec:prompt} and uses the output as the input of the next self-attention layer.
 Section~\ref{sec:combiner} discusses the details.

\textbf{\eprompt:} $\mathbf{E} = \{\ve_t\}_{t=1}^{T}$ is a set of task-dependent parameters, where $\ve_t \in \mathbb{R}^{L_e \times D}$ has a sequence length of $L_e$ and the same embedding dimension $D$ as the \gprompt, and $T$ is the total number of tasks. Different from the shared \gprompt, each $\ve_t$ is associated with a task-specific key $\vk_t \in \mathbb{R}^{D}$, which is also a learnable parameter that aims to capture representative features of a task. For an input example from the $t$-th task, to attach \eprompt to the $j$-th MSA layer, we apply the prompting function in a similar way:
%\vspace{-.5mm}
\begin{equation} \label{eq:eprompt}
    \vh_e^{(j)} = \pfunc \left(\ve_t, \vh^{(j)}\right).
    %\vspace{-1.5mm}
\end{equation}
Moreover, we update the corresponding $\vk_t$ to match the feature of the input instance via a matching loss $\mathcal{L}_{\text{match}}$, such that $\vk_t$ becomes ``closer'' to examples from the $t$-th task than other keys. At test time, inspired by the strategy proposed in \cite{wang2021learning}, we propose to adopt a query function $q$ on the test sample to search for the best match from the task keys, and select the corresponding \eprompt to use.
% See section~\ref{sec:main_obj} for specific definitions about $q$ and $\mathcal{L}_{\text{match}}$.
% $\mathcal{L}_{\text{match}}$ aims at matching an input example to a proper task key $\vk_t$, thus selecting the corresponding \eprompt, $\ve_t$. 
Although it is interesting to design various matching and query strategies by introducing additional components, it actually violates the principle of parsimony in continual learning \cite{hadsell2020embracing,wang2020learn}. Fortunately, as suggested in \cite{wang2021learning}, we can directly use the whole pre-trained model as the query function: $q(\vx) = f(\vx)[0]$ (the feature vector corresponding to \verb|[class]| token~\cite{dosovitskiy2020image}), and cosine similarity as $\gamma$. Thus, the matching loss takes the following form:
\begin{equation} \label{eq:match_loss}
    \mathcal{L}_{\text{match}}(\vx, \vk_t) = \gamma(q(\vx), \vk_t), \quad \vx \in \mathcal{D}_t.
\end{equation}
For a test example $\vx$, we simply choose the best matched task key index via ${\operatorname{argmin}}_{t}\  \gamma(q(\vx), \vk_t)$.
%, which acts like an estimation of test-time task identity that takes into account the similarity between tasks. 
%Thus, even if mismatching happens, it still ensures the selection of \eprompt from one of the most similar tasks.*h\
% In Appendix~\ref{app:perfect_match}, we study its relation to test-time id estimation.
We show the relationship between query accuracy and final performance in Appendix~\ref{app:perfect_match}.
We empirically discover this matching loss and the corresponding query mechanism works fairly well for all benchmarks.

% \begin{figure}[t]
% \centering 
% \includegraphics[width=0.9\linewidth]{figures/prompt_combiner.pdf} 
% %\vspace{-.3cm}
% % https://www.figma.com/file/y87HDSL0XmtTfDwlAppTRO/DualPrompts?node-id=2%3A3
% \caption{Overview of the \combiner function, which combines a pre-trained backbone (frozen) with multiple self-attention layers and two types of prompts. \combiner provides modularized configurations of the prompts and serves as an unified framework for prompt-based continual learning.
% %G-Prompts, E-Prompts and their associated keys are trainable variables.
% }
% \label{fig:combiner}
% \end{figure} 

% \subsection{Modularized Prompt Attaching Framework}
\subsection{Prompt attaching: where and how?}\label{sec:combiner}
%\zifeng{Re-organize this section: highlight and motivation}
G- and E-prompts encode respective type of instructions during training with the backbone and cooperatively instruct the model to make predictions at inference. We have showcased how to attach them to a single MSA layer in Section~\ref{sec:g-e-prompt}. Most existing prompt-related work simply place prompts only at the first MSA~\cite{wang2021learning,lester2021power}, or at every MSA layer~\cite{li2021prefix,liu2021p}. However,
we argue that it is crucial to explore \emph{where} and \emph{how} to attach both types of prompts. %under the continual visual learning setting. We motivate our design choices below.

\textbf{Where: Decoupled prompt positions.} %\combiner offers an option to place G- and E-prompts to different layers of the backbone model. 
Intuitively, different layers of the backbone have different levels of feature abstraction~\cite{raghu2021vision}. Therefore, when learning tasks sequentially, some layers of representations can have higher responses to task-specific knowledge than others, vise versa for task-invariant knowledge. This motivates us to give the two types of prompts more flexibility to attach to the most proper positions in a decoupled way, thus different instructions can interact with the corresponding representations more effectively. %Moreover, \combiner is configurable for multi-layered prompt integration (when $\jgp > \igp$ or $\jep > \iep$), which also provides a controllable way to enhance the representation power of the instructions.
% which performs much better single layer~\cite{lester2021power} and also all layers~\cite{li2021prefix}.

With a slight abuse of notation, we introduce the multi-layered extension of both types of prompts: $\vg = \{\vg^{(l)}\}_{l=\igp}^{\jgp}$, where $\vg^{(l)} \in \mathbb{R}^{L_g \times D}$ is the G-Prompt to be attached to the $l$-th MSA layer. We also define $\ve_t = \{\ve^{(l)}_t\}_{l=\iep}^{\jep}$ similarly. In this way, we are able to attach the \gprompt $\vg^{(l)}$ from the $\igp$-th to the $\jgp$-th MSA layers, and attach the \eprompt $\ve^{(l)}_t$ from the $\iep$-th to the $\jep$-th MSA layers. And most importantly, $(\igp, \jgp)$ and $(\iep, \jep)$ could be totally different or non-overlapping. In our experiments, we empirically search for a certain set of $\igp, \jgp, \iep, \jep$ on a validation set and discover that it performs consistently well across different benchmarks. Note that we make a simplified assumption that the chosen indices of MSA layers to attach prompts are contiguous, which already achieves state-of-the-art performance in our empirical evaluation. However, there could be more advanced ways to auto-search the configuration, which we treat as valuable future work.

% First of all, $\vg$ and $\ve_t$'s have the flexibility to be placed at different depth of the model. Since different layers of deep architectures captures different aspects of feature/knowledge \cite{XXX}, when learning from a sequence of tasks, some of them are more transferable between tasks while others may be more specialized for certain tasks. Therefore, it is natural for \gprompt and \eprompt to be placed at different layers to capture task-invariant and task-specific features respectively.

% \zizhaoz{Put those observations in experiments after the validation search.}
% Secondly, \combiner enables multi-layered prompts, instead of only prompting the very first attention layer~\cite{lester2021power}, or prompting every attention layer~\cite{li2021prefix}. Intuitively, adding prompts to multiple layers increases the learning capacity. However, in the continual learning context, we have to be more cautious to control the number of prompted layers: too many layers might lead to more forgetting for \gprompt, and also results in more additional parameters for both types of prompts.

\textbf{How: Configurable prompting function.} %\combiner introduces the prompting function $\pfunc$ to control the way we prepend prompts to the embedding features, \zifeng{which influences the underlying instruction process.} 
The prompting function $\pfunc$ controls the way we combine prompts with the embedding features. From another perspective, $\pfunc$ directly affects how the high-level instructions in prompts interact with low-level representations. Thus, we believe a well-designed prompting function is also vital for the overall continual learning performance. Although \method is  compatible with various prompting functions, here we exemplify and study two mainstream realizations in the NLP community - Prompt Tuning (Pro-T)~\cite{lester2021power} and Prefix Tuning (Pre-T)~\cite{li2021prefix}.

% The specific implementation of $\pfunc$ can be user-defined. Here we exemplify and study two mainstream realizations in the NLP community - prompt-tuning (Pro-T) \cite{lester2021power} or prefix-tuning (Pre-T) \cite{li2021prefix}. %To the best of our knowledge, we are the first to empirically study the influence of prompting function in the continual learning context. 

Specifically, applying a prompting function can be viewed as modifying the inputs of the MSA layers \cite{vaswani2017attention}. Let the input to the MSA layer be $\vh \in \mathbb{R}^{L \times D}$, and we further denote the input query, key, and values for the MSA layer to be $\vh_Q,  \vh_K, \vh_V$, respectively. Recall that the MSA layer is proposed by \cite{vaswani2017attention}:
% Proposed by Vaswani \textit{et al.,} \cite{vaswani2017attention}, the multi-head self-attention (MSA) is defined as:
\begin{equation*}
%\vspace{-1mm}
\begin{aligned}
&\operatorname{ MSA }(\vh_Q, \vh_K, \vh_V) =\operatorname { Concat }\left(\operatorname { h }_{1}, \ldots, \operatorname { h }_{\mathrm{m}}\right) W^{O} \\
&\text { where} \operatorname{h}_{\mathrm{i}} =\operatorname{Attention}\left(\vh_Q W_{i}^{Q}, \vh_K W_{i}^{K}, \vh_V W_{i}^{V}\right),
\end{aligned}
%\vspace{-1mm}
\end{equation*}
where $W^{O}$, $W_{i}^{Q}$, $W_{i}^{K}$, and $W_{i}^{V}$ are projection matrices. $m$ is the number of heads. In ViT, $\vh_Q = \vh_K = \vh_V$.
For simplicity, we define a unified prompt parameter $\vp \in \mathbb{R}^{L_{p} \times D}$ ($\vp$ could be either single-layered G or E-Prompt).

\textbf{Prompt Tuning (Pro-T)} 
%\sayna{I know you mentioned the full name above but i think it's worth mentioning it again as Prompt tuning and Prefix tuning} 
prepends prompts to the input tokens, which is equivalent to concatenate the same prompt parameter $\vp$ to $\vh_Q$, $\vh_K$, and $\vh_V$, 
\begin{equation}
f_{\operatorname{prompt}}^{\operatorname{{Pro-T}}}(\vp, \vh) = \operatorname{ MSA }([\vp;\vh_Q], [\vp;\vh_K], [\vp;\vh_V]),  
\end{equation}
where $[\cdot;\cdot]$ defines the concatenation operation along the sequence length dimension. The output length increases, resulting the output dimension as $\mathbb{R}^{(L+L_p) \times D}$. The operation is equivalent to how \verb|[class]| is added \cite{dosovitskiy2020image} at the first MSA layer. 

\textbf{Prefix Tuning (Pre-T)} splits $\vp$ into $\vp_K, \vp_V \in \mathbb{R}^{L_p/2 \times D}$, and prepends them to to $\vh_K$ and $\vh_V$ respectively, while keep $\vh_Q$ as-is:
\begin{equation}
f_{\operatorname{prompt}}^{\operatorname{{Pre-T}}}(\vp, \vh) = \operatorname{ MSA }(\vh_Q, [\vp_k;\vh_K], [\vp_v;\vh_V]).  
\end{equation}
Compared with Pro-T, the output sequence length remains the same as input $\vh \in \mathbb{R}^{L \times D}$. 
Section~\ref{sec:exp-combiner} studies both versions empirically and discusses the intuition behind their difference in performance from a continual learning perspective. 

\subsection{Overall objective for \method} \label{sec:main_obj}
The full picture of \method at training and test time is described in Algorithm~\ref{alg:dualprompt_train} and~\ref{alg:dualprompt_test}, respectively, in Appendix~\ref{app:algorithm}. Following the design patterns discussed in Section~\ref{sec:combiner}, we denote the architecture with prompts attached by $f_{\vg, \ve_t}$. Then we transform our input $\vx$ from the $t$-th task via $f_{\vg, \ve_t}$ and send it to the classification head $f_\phi$ parametrized by $\phi$ for prediction. Finally, we train both types of prompts, the task keys, as well as the newly-initialized classification head in an end-to-end fashion:
%\vspace{-2mm}
\begin{equation} \label{eq:full_loss}
    \underset{\vg, \ve_t, \vk_t, \phi}{\operatorname{min}}  \mathcal{L}(f_\phi(f_{\vg, \ve_t}(\vx)), y) + \lambda \mathcal{L}_{\text{match}} \left(\vx, \vk_t \right), \quad \vx \in \mathcal{D}_t,
    %\vspace{-2mm}
\end{equation}
where $\mathcal{L}$ is the cross-entropy loss, $\mathcal{L}_{\text{match}}$ is the matching loss defined in~\eqref{eq:match_loss}, and $\lambda$ is a scalar balancing factor. 

\section{Experiments}
%\zifeng{Add more insightful discussions, instead of just showing results.}
% We validate \method by conducting comprehensive experiments, following the settings widely used in prior works \cite{wu2019large,wang2021learning}. 
%  Our method is novel in this design principle compared with prior works, especially rehearsal-based method and architecture-based methods. 
% \zifeng{In this section, we first conduct extensive experiments with careful setups to compare fairly with previous related methods to demonstrate the superior performance of \method. We then design exploratory and ablation studies to further highlight the key designs.} \sayna{this sentence is a bit off topic here as the beginning of the experimental section}
%\zizhaoz{Reduce repeated and less informative content that is easy to know, provide more insightful prerequisites to readers.}
%We first introduce our newly proposed evaluation benchmark, Split ImageNet-R, in addition to detailed experiment settings. We then compare \method with representative continual learning methods that are compatible with the class-incremental setting, including state-of-the-art rehearsal and non-rehearsal based methods. Finally, we conduct extensive ablation studies and exploratory experiments to show the effectiveness of each component of the proposed \combiner framework.

\subsection{Evaluation benchmarks}
\textbf{Split ImageNet-R}. The Split ImageNet-R benchmark is build upon ImageNet-R \cite{hendrycks2020many} by dividing the 200 classes randomly into 10 tasks with 20 classes per task. We split the dataset into training and test set with 24,000 and 6,000 images respectively. We further sample $20\%$ from the training set as validation data for prompt attaching design search. The original ImageNet-R includes newly collected data of different styles, such as cartoon, graffiti and origami, as well as hard examples from ImageNet~\cite{deng2009imagenet} that standard models, \eg, ResNet~\cite{he2016deep}, fail to classify. We believe the Split ImageNet-R is of great importance to the continual learning community, for the following reasons: 1) Split ImageNet-R contains classes with different styles, which is closer to the complicated real-world problems. 2) The significant intra-class diversity (see Appendix~\ref{app:split_imr}) poses a great challenge for rehearsal-based methods to work effectively with a small buffer size (see Figure~\ref{fig:intro}), thus encouraging the development of more practical, rehearsal-free methods. 3) Pre-trained vision models are useful in practice for many fields \cite{ridnik2021imagenet,kolesnikov2020big}, including continual learning. However, their training set usually includes ImageNet. Thus, Split ImageNet-R serves as a relative fair and challenging benchmark, and an alternative to ImageNet-based benchmarks \cite{rebuffi2017icarl,wu2019large} for continual learning that uses pre-trained models.

% \textbf{Standard benchamrks.} We also use the widely used \textbf{Split CIFAR-100}~\cite{lopez2017gradient, chaudhry2018efficient} and \textbf{5-datasets}~\cite{ebrahimi2020adversarial, mehtaempirical}. 
\textbf{Split CIFAR-100.} Split CIFAR-100 is a widely-used benchmark in continual learning literature. It splits the original CIFAR-100 \cite{krizhevsky2009learning} into 10 disjoint tasks, with 10 classes per task. Although it is a relatively simple task for image classification under the i.i.d. setting, it sufficiently makes advanced CL methods expose large forgetting rate in class-incremental learning.

% \textbf{5-datasets.} 5-datasets is a collection of five diverse image classification datasets, CIFAR-10~\cite{krizhevsky2009learning}, MNIST~\cite{lecun1998mnist}, Fashion-MNIST~\cite{xiao2017fashion}, SVHN~\cite{netzer2011reading}, and notMNIST~\cite{notmnist}. Despite the simplicity of each task in 5-datasets, the benchmark mimics the real-world setting where task diversity is large, thus contributing to a more comprehensive evaluation of CL methods.

We use Split ImageNet-R and Split CIFAR-100 to demonstrate our main results in Section~\ref{sec:exp-class-inc}, and additionally conduct experiments on 5-datasets for completeness in the Appendix~\ref{app:5-datasets}. 

% \zifeng{We specifically choose Split CIFAR-100 since almost all continual learning methods treat it as a standard and challenging class-incremental learning benchmark, and we can reliably reproduce the result to provide fair comparison. 5-datasets is also well-recognized~\cite{ebrahimi2020adversarial,mehtaempirical,wang2021learning}, but most importantly, it mimics the real-world setting where task diversity is large, thus contributing to a more comprehensive evaluation of CL methods.} 
% \subsection{Experimental details and evaluation metrics.}
% \zifeng{Shorten them into one sentence in 5.3.}\\
% \noindent\textbf{Experimental details.} Please refer to Appendix~\ref{app:exp_details}.\\
% \noindent\textbf{Evaluation Metrics.}
% we use the widely-used \emph{Average accuracy} (higher is better) and \emph{Forgetting} (lower is better)~\cite{lopez2017gradient,chaudhry2018efficient,mai2021online} as our evaluation metrics. The exact definition of both metrics are shown in the Appendix~\ref{app:eval_metrics}. 
% Note that we use {Average accuracy} as the major criteria for model performance, since it reflects both the learning capacity of the model, and the model's robustness to catastrophic forgetting. \sayna{not exactly - a model can have a high avg accuracy but a very bad forgetting} \zifeng{discuss it}

\begin{table}[t!]
%\vspace{-2mm}
\caption{\smaller Results on class-incremental learning (i.e., task identity is unknown at test time). We compare and group methods by buffer sizes. 0 means no rehearsal is used, when most SOTA methods are not applicable anymore. Note that the chosen buffer sizes here are considered sufficiently large sizes that are used in prior works for Split CIFAR-100 \cite{chaudhry2018efficient,buzzega2020dark}.
They are large enough even for training a supervised counterpart -- e.g. GDumb~\cite{prabhu2020gdumb} trains on the i.i.d sampled buffer with this size and demonstrated competitive results, making continual training unnecessary. 
%\zizhaoz{rank the rows by either year or accuracy.}
}
\label{table:cifar_imagenet}
\begin{center}
\scalebox{0.8}{
\begin{tabular}{l||c|cc||c|cc}
\toprule 
 \multirow{2}{*}{\textbf{Method}} & \multirow{2}{*}{\textbf{Buffer size}} & \multicolumn{2}{c||}{\textbf{Split CIFAR-100}} & \multirow{2}{*}{\textbf{Buffer size}} & \multicolumn{2}{c}{\textbf{Split ImageNet-R}} \\
%  &
& &  Avg. Acc ($\uparrow$) & Forgetting ($\downarrow$) & & Avg. Acc ($\uparrow$) & Forgetting ($\downarrow$) \\
\midrule
   ER \cite{chaudhry2019tiny} & \multirow{6}{*}{1000} & 67.87\scriptsize{$\pm$0.57} & 33.33\scriptsize{$\pm$1.28} & \multirow{6}{*}{1000} & {55.13\scriptsize{$\pm$1.29}} & 35.38\scriptsize{$\pm$0.52} \\
   BiC \cite{wu2019large} & & 66.11\scriptsize{$\pm$1.76} & 35.24\scriptsize{$\pm$1.64} && 52.14\scriptsize{$\pm$1.08} & 36.70\scriptsize{$\pm$1.05} \\
  GDumb \cite{prabhu2020gdumb} & & 67.14\scriptsize{$\pm$0.37} & - && 38.32\scriptsize{$\pm$0.55} & - \\
 DER++ \cite{buzzega2020dark} & & 61.06\scriptsize{$\pm$0.87} & 39.87\scriptsize{$\pm$0.99} && 55.47\scriptsize{$\pm$1.31} & 34.64\scriptsize{$\pm$1.50} \\
 Co$^2$L \cite{cha2021co2l} & & 72.15\scriptsize{$\pm$1.32} & 28.55\scriptsize{$\pm$1.56} && 53.45\scriptsize{$\pm$1.55} & 37.30\scriptsize{$\pm$1.81} \\
 % L2P-R \cite{wang2021learning} & & {84.21\scriptsize{$\pm$0.53}} & {7.72\scriptsize{$\pm$0.77}} && {85.56\scriptsize{$\pm$0.95}} & {4.22\scriptsize{$\pm$0.03}} \\
 \midrule
 ER \cite{chaudhry2019tiny} &\multirow{6}{*}{5000 }& 82.53\scriptsize{$\pm$0.17} & 16.46\scriptsize{$\pm$0.25} &\multirow{6}{*}{5000 }& 65.18\scriptsize{$\pm$0.40} & 23.31\scriptsize{$\pm$0.89} \\
 BiC \cite{wu2019large} & & 81.42\scriptsize{$\pm$0.85} & 17.31\scriptsize{$\pm$1.02} && 64.63\scriptsize{$\pm$1.27} & 22.25\scriptsize{$\pm$1.73} \\
 GDumb \cite{prabhu2020gdumb} & & 81.67\scriptsize{$\pm$0.02} & - && 65.90\scriptsize{$\pm$0.28} & -  \\
 DER++ \cite{buzzega2020dark} & & 83.94\scriptsize{$\pm$0.34} & 14.55\scriptsize{$\pm$0.73} && 66.73\scriptsize{$\pm$0.87} & 20.67\scriptsize{$\pm$1.24} \\
 Co$^2$L \cite{cha2021co2l} & & 82.49\scriptsize{$\pm$0.89} & 17.48\scriptsize{$\pm$1.80} && 65.90\scriptsize{$\pm$0.14} & 23.36\scriptsize{$\pm$0.71} \\
 \midrule
 % FT-seq-frozen & \multirow{6}{*}{0} & 17.72\scriptsize{$\pm$0.34} & 59.09\scriptsize{$\pm$0.25} & \multirow{6}{*}{0} & 39.49\scriptsize{$\pm$0.12} & 42.62\scriptsize{$\pm$0.20} \\ 
 FT-seq & \multirow{5}{*}{0} & 33.61\scriptsize{$\pm$0.85} & 86.87\scriptsize{$\pm$0.20} & \multirow{5}{*}{0} & 28.87\scriptsize{$\pm$1.36} & 63.80\scriptsize{$\pm$1.50}\\
 EWC \cite{kirkpatrick2017overcoming} & & 47.01\scriptsize{$\pm$0.29} & 33.27\scriptsize{$\pm$1.17} && 35.00\scriptsize{$\pm$0.43} & 56.16\scriptsize{$\pm$0.88} \\
 LwF \cite{li2017learning} & & 60.69\scriptsize{$\pm$0.63} & 27.77\scriptsize{$\pm$2.17} && 38.54\scriptsize{$\pm$1.23} & 52.37\scriptsize{$\pm$0.64} \\
 {L2P} \cite{wang2021learning} & & {83.86\scriptsize{$\pm$0.28}} & {7.35\scriptsize{$\pm$0.38}} &&{61.57\scriptsize{$\pm$0.66}} & {9.73\scriptsize{$\pm$0.47}} \\
 \bf{\method} & & \bf 86.51\scriptsize{$\pm$0.33} & \bf 5.16\scriptsize{$\pm$0.09} & & \bf 68.13\scriptsize{$\pm$0.49} & \bf 4.68\scriptsize{$\pm$0.20}  \\
\midrule
Upper-bound & -& 90.85\scriptsize{$\pm$0.12} & - & - & 79.13\scriptsize{$\pm$0.18} & - \\
\bottomrule
\end{tabular}
}
\end{center}%\vspace{-.6cm}
\end{table} 

\subsection{Comparison with state-of-the-arts} \label{sec:exp-class-inc}
We compare \method against representative baselines and state-of-the-art methods. 
Please refer to Appendix~\ref{app:exp_details} for experimental details. 
We use the widely-used \emph{Average accuracy} (higher is better) and \emph{Forgetting} (lower is better)~\cite{lopez2017gradient,chaudhry2018efficient,mai2021online} as our evaluation metrics. The definitions of both metrics are in Appendix~\ref{app:eval_metrics}.

To make the comparison fair and precise, we first compare \method with regularization-, rehearsal- and prompt-based methods, which are compatible with transformer-based models, in Table~\ref{table:cifar_imagenet}. We then compare \method with architecture-based methods, which are mostly compatible with ConvNets, using a different protocol in Table~\ref{table:architecture}. 
% We leave comparison with architecture-based methods in Section~\ref{sec:exp-architecture}.
%\sayna{why? you are basically saying you think that result is not one of the main results and discredit that set of experiment while some folks out there do not think that way. I think now that you have the full set of experiments with comparison with all types, we should make section 5.4 as a part of the current section } 

% \zizhaoz
\begin{itemize}
\item \textbf{Comparing methods.} We select representative methods including \textsf{EWC} \cite{kirkpatrick2017overcoming}, \textsf{LwF} \cite{li2017learning}, \textsf{ER} \cite{chaudhry2019tiny,hayes2019memory}, \textsf{GDumb} \cite{prabhu2020gdumb}, \textsf{BiC} \cite{wu2019large}, \textsf{DER++} \cite{buzzega2020dark}, \textsf{Co$^2$L} \cite{cha2021co2l} and \textsf{L2P}~\cite{wang2021learning}, from all categories. Please see Appendix~\ref{app:comparing_methods} for details.
    
    \item \textbf{Naive baselines.} For better demonstration of the relative effectiveness of all methods, we also include: \textsf{FT-seq}, the naive sequential training, and \textsf{Upper-bound}, the usual supervised finetuning on the i.i.d.~data of all tasks.
\end{itemize}

% Moved to appendix
% To ensure fair comparison, every aforementioned methods start from the same ImageNet pre-trained ViT-B/16 \cite{vit}, following the setting in~\cite{wang2021learning}. We carefully re-implement these method and use hyper-parameters by referring to their original source code. Moreover, we make the pre-trained model fully trainable for all methods (except L2P and \method), as we empirically observe they could not learn as good with a frozen backbone due to limited learning capacity. 

%\hl{Note that we do not include several interesting related works since they are either limited by their usage on simpler task-incremental setting} \zizhaoz{Not clear enough about "limited by their usage".} \cite{ke2020continual,ebrahimi2020adversarial,pham2020contextual}, or specific architecture design~\cite{pham2021dualnet}.

%\subsection{Results on continual learning benchmarks} 
Table~\ref{table:cifar_imagenet} reports the performance of all comparing methods on Split CIFAR-100 and Split ImageNet-R. Our proposed method, \method, outperforms all methods consistently, including non-rehearsal based methods and rehearsal-based methods with a large buffer size. 
%Note that since each task in 5-datasets is relatively easy, rehearsal-based methods naturally requires less examples to perform well, however, a buffer size of 500 is already considered large enough in prior work~\cite{mehtaempirical, ebrahimi2020adversarial}. 
When the buffer size is 5000 (10\% of the CIFAR-100 training set and $>$20\% of the ImageNet-R training set), all rehearsal-based methods are fairly close to GDumb, indicating that performing rehearsal-based continual learning likely provide no performance gain than supervised training on the buffered data as GDumb does. \textit{\method achieves better performance without any buffered data}. 
Moreover, from Table~\ref{table:cifar_imagenet}, as well as Figure~\ref{fig:intro}, we can observe the performance of rehearsal-based methods drops sharply when the buffer size shrinks. This again suggests the clear advantage of \method as a rehearsal-free method. For the non-rehearsal based methods, only L2P performs close to our methods. Nevertheless, \method still beats L2P significantly by a 3\%-7\% margin on Average accuracy, thanks to our novel design of the two complementary prompts, which successfully reduces catastrophic forgetting.

\begin{table*}[t!]
\small
%\vspace{-2mm}
\caption{\smaller Comparison with architecture-based methods on Split CIFAR-100. We use \texttt{Diff = Upper-Bound Acc - Method Acc} (lower is better), to measure how close the performance to the upper-bound of the used backbone.}
\label{table:architecture}
\begin{minipage}{0.98\textwidth}
\begin{center}
\scalebox{0.8}{
\begin{tabular}{l|c|lc|c|>{\centering\arraybackslash}p{1.8cm}>{\centering\arraybackslash}p{1.8cm}}
\toprule 
 \multirow{2}{*}{\textbf{Method}} & \multirow{2}{*}{\textbf{Backbone}} & \multirow{2}{*}{\textbf{Avg. Acc ($\uparrow$)}} & \multirow{2}{*}{\textbf{Diff ($\downarrow$)}} & \multirow{2}{*}{\textbf{Buffer size}} &  \multicolumn{2}{c}{\textbf{Additional Parameters}} \\
& &  & & & MB & \% \\
\midrule
  Upper-bound & \multirow{5}{*}{ResNet18}& 80.41$^\dagger$ & - & - & - & - \\
   SupSup~\cite{wortsman2020supermasks} & & 28.34\scriptsize{$\pm$2.45}$^\ddagger$ & 52.07 & 0 & 3.0 & 6.5\% \\
   DualNet~\cite{pham2021dualnet} & & 40.14\scriptsize{$\pm$1.64}$^\ddagger$ & 40.27 & 1000 & 5.04 & 10.9\% \\
  RPSNet~\cite{rajasegaran2019random} & & 68.60$^\dagger$ & 11.81 & 2000 & 181 & 404\% \\
 DynaER~\cite{yan2021dynamically} & & 74.64$^\dagger$ & 5.77 & 2000 & 19.8 & 43.8\% \\
%  \midrule
%  DynaER~\cite{yan2021dynamically} & ResNet50 & -$^\ddagger$ & - & 2000 & - & -\% \\
 \midrule
 Upper-bound & \multirow{2}{*}{ResNet152}& 88.54$^\dagger$ & - & - & - & - \\
 DynaER~\cite{yan2021dynamically} & & 71.01\scriptsize{$\pm$0.58}$^\ddagger$ & 17.53 & 2000 & 159 & 68.5\% \\
 % L2P-R \cite{wang2021learning} & & {84.21\scriptsize{$\pm$0.53}} & {7.72\scriptsize{$\pm$0.77}} && {85.56\scriptsize{$\pm$0.95}} & {4.22\scriptsize{$\pm$0.03}} \\
 \midrule
 % FT-seq-frozen & \multirow{6}{*}{0} & 17.72\scriptsize{$\pm$0.34} & 59.09\scriptsize{$\pm$0.25} & \multirow{6}{*}{0} & 39.49\scriptsize{$\pm$0.12} & 42.62\scriptsize{$\pm$0.20} \\ 
Upper-bound & \multirow{3}{*}{ViT-B/16}& 90.85\scriptsize{$\pm$0.12}$^\ddagger$ & - & - & - & - \\
%  {L2P-S} \cite{wang2021learning} &  & {83.62\scriptsize{$\pm$0.00}}$^\ddagger$ & 7.23 & 0 & 0.7 & 0.2\% \\
 {L2P} \cite{wang2021learning} &  & {83.86\scriptsize{$\pm$0.28}}$^\ddagger$ & 6.99 & 0 & 1.94 & 0.56\% \\
%  \bf{\method-SL} & & \bf 85.75\scriptsize{$\pm$0.00} & \bf 5.10 & 0 & \bf 0.7 &  \bf 0.2\%  \\
 \bf{\method} & & \bf 86.51\scriptsize{$\pm$0.33} & \bf 4.34 & 0 & \bf 1.90 &  \bf 0.55\%  \\
\bottomrule
\end{tabular}
}
\\
\scriptsize{$^\dagger$Reported by the original papers. $^\ddagger$ Reproduced using their original codebases.}
\end{center}
\end{minipage}
%\vspace{-.5cm}
\end{table*}

\textbf{Architecture-based methods.} We compare against representative class-incremental learning methods, including DualNet~\cite{pham2021dualnet}, SupSup~\cite{wortsman2020supermasks}, DynaER~\cite{yan2021dynamically} and RPSNet~\cite{rajasegaran2019random}. Please see Appendix~\ref{app:comparing_methods} for details.

Prior architecture-based methods, which are based on ConvNet, are not trivial to migrate to transformer-based models. Moreover, different architecture-based methods usually add different amount of additional parameters. %In addition, the majority of architecture-based methods only focus on the easier task-incremental learning setting with given test-time task identity \cite{}. \zizhaoz{Seems not relevant content.}
To enable a relative fair comparison between these methods, we introduce a metric to measure how close the performance of a certain method is to the upper-bound performance, \ie trained under the i.i.d. setting, of a given architecture.
Table~\ref{table:architecture} shows the results on Split-CIFAR100. \method achieves the best accuracy and its difference to upper-bound is only $4.34\%$, with minimal additional parameters and no buffer. 
%(\zifeng{note that the single-layered version of \method even has fewer parameters, \ie, $0.19\%$ of the full model, but remains competitive result, see Section~\ref{sec:ablation}}).
The strongest competitor on ResNet, DynaER, on the contrary, requires a buffer of $2,000$ images and include $43.8\%$ additional parameters. 

%\vspace{-1.5mm}
\subsection{Does stronger backbones naively improve CL?} 
%\vspace{-1.5mm}
Our method builds upon more advanced yet bigger backbones than many previous methods. We think understand this question is very important for fair comparison and future research. 
Although pre-trained ViT is a stronger backbone than common ConvNets, it is not necessarily translate to continual learning performance. The observations we shown here are similar to what reported in a very recent study about how large architecture help continual learning~\cite{mirzadeh2022architecture}. First, this fact can be seen from Table \ref{table:cifar_imagenet}, where well-known general methods still suffer large forgetting rate given this backbone.  %
We have tried to use ImageNet pre-trained ResNet for competing methods in Table~\ref{table:architecture}, which leads to no improvement upon the reported numbers in Table~\ref{table:architecture}. This further indicates that a pre-trained model is not a ``panacea'' for continual learning without being leveraged properly.
We also equip best-performing DynaER~\cite{yan2021dynamically} with ImageNet pre-trained ResNet152 (60M parameters), which has close upper-bound performance to ViT-B/16 (86M). DynaER learns weight masks, a popular strategy for architecture-based methods \cite{wortsman2020supermasks}, to dynamically expand architectures. 
However, results in Table~\ref{table:architecture} show worse performance (we sweep their suggested hyper-parameters to report the best performance), a similar observation is shown in their original paper when scaling DynaER to ResNet32. That being said, how to effectively utilize large models under traditional architecture-based methods remains an open question. \method is novel at wisely leveraging the state-of-the-art vision backbones to solve challenges in continual learning. 
%\sayna{I liked this section a lot! :) }

%This observation indicates that utilizing pre-trained model 
%is not a shortcut to approach better performance in the continual learning scenario. The original paper in DynaER also find.
% }

% \zizhaoz{I worry the below, with only Dyna is not strong enough to claim such thing. let me do a rehearse.}
% To further highlight that the scale of model is not the reason of our superior performance, we provide DynaER with the larger-scale, ImageNet pre-trained ResNet152, which has a very close upper-bound performance to the ViT-B/16 model we use. We make sure the model converges well when train on a single task. However, DynaER got even worse performance, both in terms of average accuracy and difference to upper-bound. This observation indicates that although larger scale pre-trained model may leads to better upper-bound performance, it is not a shortcut to approach better performance in the continual learning scenario.

\begin{figure}[t]
\centering
\begin{minipage}{0.48\textwidth}
  \centering
  \includegraphics[width=.9\textwidth]{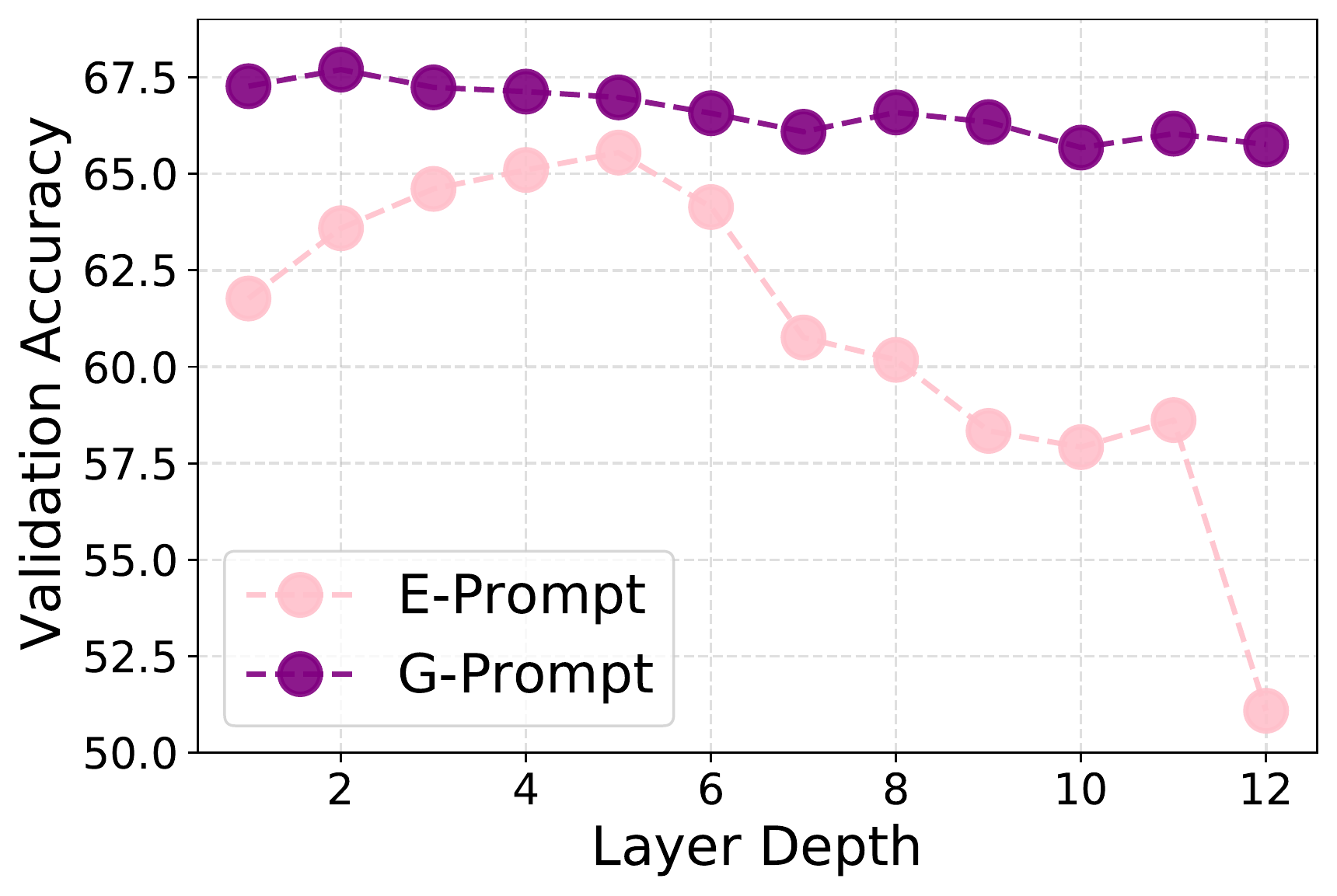}
  %\vspace{-.3cm}
  \captionof{figure}{\smaller Effects of position to attach prompts on Split ImageNet-R validation set. We empirically observe that attaching G- and E-Prompts to the 2nd and 5th MSA layer results in the best performance.}
  \label{fig:prompt_depth}
\end{minipage}
\hfill
\begin{minipage}{0.48\textwidth}
\centering
  \includegraphics[width=.9\textwidth]{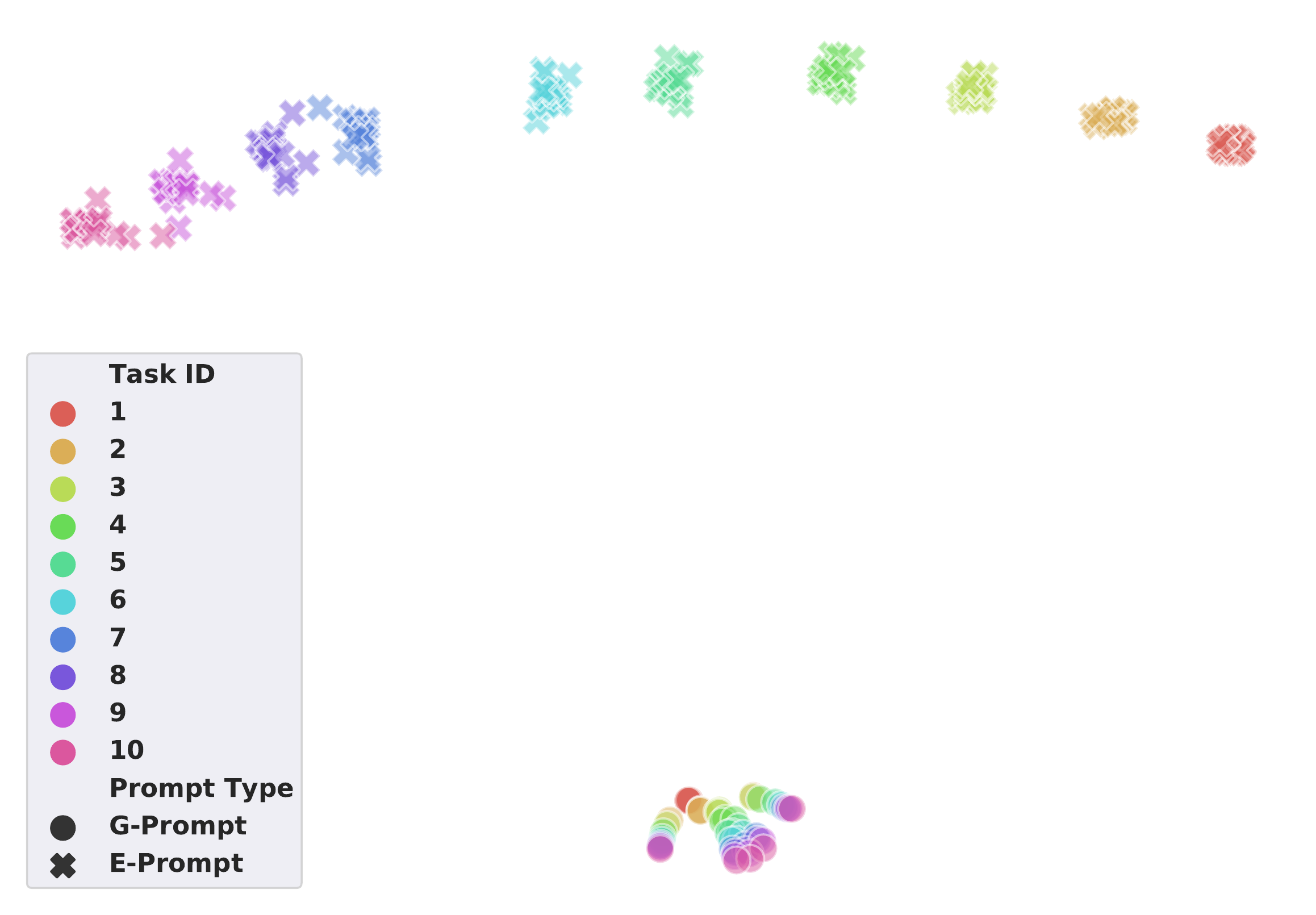}
  %\vspace{-.2cm}
  \captionof{figure}{\smaller t-SNE visualization of G- and E-prompts. Each point represents a prompt vector of dimension $768$. \eprompt{s} are taken from the final model, while \gprompt{s} are taken from model snapshots after trained on each task.}
  %We can observe that \eprompt{s} are well-separated, indicating they are learning task-specific knowledge. Meanwhile, the \gprompt{s} are quite centered and only differ slightly between tasks, which suggests they are learning transferable knowledge.}
  \label{fig:prompt_visualization}
\end{minipage}
%\vspace{-.5cm}
\end{figure}

%\vspace{-1.5mm}
\subsection{Exploration of where and how to attach prompts}
%\vspace{-1.5mm}
\label{sec:exp-combiner}
We have shown the best performing model of \method in Section~\ref{sec:exp-class-inc}. In this section, we explore \textit{where and how} to attach prompts and enhance their influences to the overall performance. We also present critical empirical observations that lead to interesting future research.

\noindent
\textbf{Position of prompts.}
To explore the most proper position to insert the \gprompt and \eprompt, we use a heuristic search strategy on the validation set of Split ImageNet-R. We first set $\iep = \jep$, \ie, only insert \eprompt at a single MSA layer. The lower line of Figure~\ref{fig:prompt_depth} shows that placing \eprompt at the 5th MSA layer leads to the best performance. We then search extend \eprompt to multi-layer and found that $\iep = 3, \jep = 5$ performs the best (see Appendix~\ref{app:multi_layer}).
We then study the case of $\igp = \jgp$ based on the optimal setting of \eprompt. Interestingly, the upper line of Figure~\ref{fig:prompt_depth} shows that placing \gprompt at the 2nd MSA layer leads to the best performance. We also extend \gprompt into its multi-layered counterparts and conduct searching experiments and find the best choice to be $\igp = 1, \jgp = 2$ (see Appendix~\ref{app:multi_layer}).

Interestingly, we observe that the best depth are different, and the final layers to attach G- and E-Prompts are non-overlapping. In particular, $\igp > \iep$, which suggests that \gprompt captures task-invariant knowledge better at shallower layer, while \eprompt captures task-specific knowledge better at deeper layer. This observation also fits the intuition that different layers in deep learning models capture different types of knowledge~\cite{raghu2021vision,zeiler2014visualizing}, and thus naturally fit different prompts. This also justifies decoupling positions of G- and E-Prompts as a reasonable option. Moreover, %both \eprompt and \gprompt exhibit the lowest performance when placed at the deepest layers. 
when attaching them to top layers, both \eprompt and \gprompt exhibit the worst performance. We speculate prompts need to be attached to shallower layers in order to condition more layers of the pre-trained model and thereby offer effective instructions.

% in order to interact with more layers of pre-trained features and thereby offer effective instructions, otherwise shallower layers will be computed independent of the prompts.

% will have less influence on the feature extraction process, since shallower layers will be computed independent of the prompts.}

% We suspect that at deepest layers, both prompts fail to interact with the pre-trained features at shallower layers, thus their learning capacity are limited.

% Although our searching strategy does not guarantee the optimal solution for the prompt positions, we empirically find our solution works well on all benchmark datasets. We believe an automated way of deciding the optimal positions for prompts will be an interesting research topic in both transfer learning and continual learning.

% \subsubsection{How the length of prompts matters?}
% Address how to choose prompt length via grid search on the validation dataset.

% \begin{table*}[t]
% \small
% \caption{Comparison between Prompt-tuning version and Prefix-tuning version.}
% \label{table:prompt_vs_prefix}
% \begin{center}
% \scalebox{0.8}{
% \begin{tabular}{c|cc|cc}
% \toprule 
%  \multirow{2}{*}{\textbf{Prompting function}} &  \multicolumn{2}{c|}{\textbf{Split CIFAR-100}} &  \multicolumn{2}{c}{\textbf{Split ImageNet-R}}\\
% %  &
% &  Avg. Acc ($\uparrow$) & Forgetting ($\downarrow$) & Avg. Acc ($\uparrow$) & Forgetting ($\downarrow$) \\
% \midrule
% Pro-T & 83.81 & 5.94 & 64.99 & 6.81 \\
% Pre-T & 86.51 & 5.16 & 68.13 & 5.52\\

% \bottomrule
% \end{tabular}
% }
% \end{center}
% \end{table*}

\begin{table}[t]
    \begin{minipage}{.48\textwidth}
        \centering 
        %\vspace{-1.cm}
        \captionof{table}{\smaller Comparison of different prompting functions: Prompt Tuning (Pro-T) v.s. Prefix Tuning (Pre-T).}
        %\vspace{0.1cm}
            \scalebox{0.8}{
            \begin{tabular}{ >{\centering\arraybackslash}p{1.8cm}|>{\centering\arraybackslash}p{2.2cm}| >{\centering\arraybackslash}p{1.4cm}| >{\centering\arraybackslash}p{1.4cm}}
            \toprule 
             \multicolumn{2}{c|}{\textbf{Prompting function}} & Pro-T & Pre-T \\
             \midrule
             Split &  Avg. Acc ($\uparrow$) & 83.81 & \bf86.51 \\
             CIFAR-100 & Forgetting ($\downarrow$) & 5.94 & \bf5.16 \\
             \midrule
             Split &  Avg. Acc ($\uparrow$) & 64.99 & \bf68.13 \\
             ImageNet-R & Forgetting ($\downarrow$) & 6.81 & \bf4.68 \\
            \bottomrule
            \end{tabular}
            }
            \label{table:prompt_vs_prefix}
    \end{minipage}
    \hfill
    \begin{minipage}{.48\textwidth}
    %\vspace{-0.45cm}
    \captionof{table}{\smaller Ablation study on Split ImageNet-R. ML means multi-layered.}
    %\vspace{0.05cm}
        \centering 
        \scalebox{0.8}{
        \begin{tabular}{ccc|cc}
        \toprule 
         \multirow{2}{*}{\textbf{G-P}}  & \multirow{2}{*}{\textbf{E-P}} & \multirow{2}{*}{\textbf{ML}} &  \multicolumn{2}{c}{\textbf{Split ImageNet-R}} \\
        %  &
        & & &  Avg. Acc ($\uparrow$) & Forgetting ($\downarrow$)  \\
        \midrule
        & & & {27.01} & {7.57} \\
        \checkmark & & & 63.41 & 6.52 \\
        & \checkmark & & 65.10 & 5.52 \\
        \checkmark & \checkmark & & 66.77 & 5.74  \\
        \checkmark & & \checkmark & 63.85 & 7.50 \\
        & \checkmark & \checkmark & 66.91 & 4.77 \\
        \checkmark & \checkmark & \checkmark & \bf68.13 & \bf 4.68\\
        \bottomrule
        \end{tabular}
        }
        \label{table:ablation}
    \end{minipage}
%\vspace{-0.6cm}
\end{table}

% \begin{table*}[t]
% \small
% \caption{Comparison between Prompt-tuning version and Prefix-tuning version.}
% \label{table:prompt_vs_prefix}
% \begin{center}
% \scalebox{0.8}{
% \begin{tabular}{c|c|cc}
% \toprule 
%  \multicolumn{2}{c|}{\textbf{Prompting function}} & Pro-T & Pre-T \\
%  \midrule
%  Split &  Avg. Acc ($\uparrow$) & 83.81 & \bf86.51 \\
%  CIFAR-100 & Forgetting ($\downarrow$) & 5.94 & \bf5.16 \\
%  \midrule
%  Split &  Avg. Acc ($\uparrow$) & 64.99 & \bf68.13 \\
%  ImageNet-R & Forgetting ($\downarrow$) & 6.81 & \bf5.52 \\
% \bottomrule
% \end{tabular}
% }
% \end{center}
% \end{table*}

\noindent
\textbf{Prompting function: Prompt v.s. Prefix.}
We further study the role of prompting function on Split CIFAR-100 and Split ImageNet-R. In prior prompt-based CL work, L2P, only Pro-T is applied without further investigation. In Table~\ref{table:prompt_vs_prefix}, we observe that Pre-T version leads to a better performance on both datasets. 
% \zizhaoz{I feel this conjuction is interesting but does not put much values in continual learning context. Maybe move to Appendix.}
% \hl{There are two intuitions to explain the observation: 1) Pre-T attaches different parameters to the query and key of the self-attention input, while Pro-T attaches the same parameters to all inputs. Thus, Pre-T has more flexibility to learn a better representation on the downstream continual learning tasks. 2) Pre-T remains the output sequence length as the original inputs, while Pro-T extend the sequence length linearly with the number of prompted layers. We hypothesize that longer sequence length may cause extra difficulty for the model to converge well, since it breaks the structure of the original pre-trained model.} 
% The comparison between Pro-T and Pre-T remains an open question in NLP~\cite{liu2021p}.
Besides its empirically better performance, Pre-T is actually more scalable and efficient when attached to multiple layers, since it results in unchanged sequence length. Nevertheless, prompting function is a flexible component of our method, and designing better prompting function is also an open research question, so we can easily plug-in any newly proposed prompting function to \method and evaluate its effectiveness on given continual learning tasks. 
\subsection{Ablation study}
%\vspace{-.1cm}
\label{sec:ablation}
% These two might go to supplement
Based on the optimal parameters searched in the previous section, we present the ablation study results in Table~\ref{table:ablation} to show the importance of each component of \method on Split ImageNet-R. Note that G-P (\gprompt) and E-P (\eprompt) alone represent the optimal single-layered version for each type of prompts ($\igp=\jgp=2, \iep=\jep=5$), while ML represents the optimal multi-layered version ($\igp=1, \jgp=2, \iep=3, \jep=5$). When all components are absent, we simply have a naive baseline with a frozen pre-trained backbone and trainable classification head.

In general, all components contribute to the final performance. 
%Another observation is that only using \gprompt provides reasonable performance on Split ImageNet-R, while performs poorly on 5-datasets. This is due to that fact that 5-datasets, though consists of simpler tasks, has greater inter-task diversity than Split-ImageNet-R. Thus only using \gprompt between tasks causes knowledge interference for 5-datasets, leading to significant forgetting.
Interestingly, adding a single-layered \gprompt alone brings substantial improvement upon the baseline, indicating that the task-invariant knowledge obtained by \gprompt generalizes pretty well across tasks. However, simply sharing knowledge between tasks introduces inevitable forgetting, due to the fact that task-specific knowledge is not properly decoupled. Thus, \eprompt alone consistently outperforms \gprompt alone, since \eprompt mitigates forgetting by separating knowledge learned from different tasks. However, only applying \eprompt ignores the task-invariant knowledge, which helps to learn future tasks. Thus, when adding \gprompt and \eprompt together to the backbone, it further enhances the overall performance by selectively decoupling the task-invariant knowledge into \gprompt and task-specific knowledge into \eprompt. We also observe that extending both prompts to its multi-layered counterparts helps consistently in all cases, due to the fact that properly adding more prompt parameters through different layers offers more representation power. 
% updated results, no longer a problem. Note that introducing \gprompt slightly increases forgetting, but its overall effect on increasing the average accuracy outweigh this negative effect (since average accuracy is our final goal for continual learning). This observation also accords with the generalization-forgetting trade-off in continual learning~\cite{raghavan2021formalizing}. 

% \begin{figure}[t]
% \centering 
% \includegraphics[width=0.80\linewidth]{figures/prompt_visualization.pdf} 
% \caption{t-SNE visualization of G- and E-prompts. Each point represents a prompt of token length 1 and hidden dimension 768. Note that the \eprompt{s} are taken from the final model after training on the sequence of all tasks, while the \gprompt{s} are taken from different model snapshots after training on each task. We can observe that \eprompt{s} are well-separated, indicating they are learning task-specific knowledge. Meanwhile, the \gprompt{s} are quite centered and only differ slightly between tasks, which suggests they are learning transferable knowledge.}
% \label{fig:prompt_visualization} 
% \end{figure}

\textbf{Visualization of G- and E-prompts.} To further understand different types of instructions learned within G- and E-prompts, we visualize these two types of prompts using t-SNE~\cite{van2008visualizing} in Figure~\ref{fig:prompt_visualization}. For a prompt with shape $L\times D$, we treat it as $L$ prompts with dimension $D$. \eprompt{s} are taken from the final model after trained on the sequence of all tasks, while the \gprompt{s} are taken from different model snapshots after training on each task. We can observe that \eprompt{s} are well-separated, indicating they are learning task-specific knowledge. Meanwhile, the \gprompt{s} are quite centered and only differ slightly between tasks, which suggests they are learning task-invariant knowledge. 

%\vspace{-.1cm}
\section{Conclusion}
%\vspace{-1.5mm}
In this paper, we present a novel method, \method, that achieves rehearsal-free continual learning under the challenging class-incremental setting. %\method decouples small learnable prompt parameters into \gprompt and \eprompt, to learn task-invariant and task-specific instructions, respectively, under a modularized prompt attaching framework.
\method presents a novel way to attach complementary prompts to a pre-trained model to learn decoupled knowledge. To comprehensively validate the proposed method, we propose a new continual learning benchmark, Split ImageNet-R, besides study on the widely-used benchmarks.
\method sets state-of-the-art performance in all metrics, surprisingly needs much lower additional memory compared with previous architecture-based and rehearsal-based methods. Empirical investigations are conducted to understand the inner-workings.
Since large-scale pre-trained models are widely used in practice for their great representation power, we believe \method serves as a starting point for real-world rehearsal-free continual learning systems. Moreover, we recommend \method as a unified framework for future prompt-based continual learning research, for its simplicity, flexibility, and strong performance.

\clearpage

\newpage
\appendix
\
\begin{algorithm}[t]
\SetAlgoLined
\textbf{Input:} Pre-trained transformer-based backbone $f$, final classification layer $f_{\phi}$, number of tasks $T$, training set ${\{(\vx_{i, t}, y_{i, t})\}_{i=1}^{n_t}}\}_{t=1}^{T}$, \gprompt $\vg$, \eprompt $\mathbf{E} = \{\ve_t\}_{t=1}^{T}$, task keys $\mathbf{K} = \{\vk_t\}_{t=1}^{T}$, $\igp, \jgp, \iep, \jep$, prompting function $f_{\text{prompt}}$, number of training epochs of the $t$-th task $M_t$ \\
\textbf{Initialize:} $\phi,\ \vg,\ \mathbf{E},\ \mathbf{K}$\\
%  \textbf{Initialization:} initialize $\theta$ by training on $\mathcal{D}_l$\; initialize $U$ by spectral embedding of $f_\theta(X)$\\
\For{$t = 1,\cdots,T$}{
    Select the task-specific \eprompt $\ve_t$ and corresponding task key $\vk_t$ \\
    Generate the prompted architecture $f_{\vg, \ve_t}$: attach $\vg$ and $\ve_t$ to $\igp$-th to $\jgp$-th and $\iep$-th to $\jep$-th MSA layers respectively, with $f_{\text{prompt}}$. \\
    \For{$e = 1, \cdots, M_t$}{
      Draw a mini-batch $B=\{ (\vx_{i, t}, y_{i, t})\}_{i=1}^{l}$ \\
      \For{$(\vx, y)$ in $B$}{
        Calculate the prompted feature by $f_{\vg, \ve_t}(\vx)$ \\
        Calculate the per sample loss $\mathcal{L}_x$ via \eqref{eq:full_loss}
      }
      Update  $\phi,\ \vg,\ \mathbf{E},\ \mathbf{K}$ by backpropagation 
    }
 }
 \caption{\method at training time}
 \label{alg:dualprompt_train}
\end{algorithm} \vspace{-.2cm}

\vspace{-.3cm}
\begin{algorithm}[t]
\SetAlgoLined
\textbf{Given components:} Pre-trained transformer-based backbone $f$, trained classification layer $f_{\phi}$, \gprompt $\vg$, \eprompt $\mathbf{E} = \{\ve_t\}_{t=1}^{T}$, task keys $\mathbf{K} = \{\vk_t\}_{t=1}^{T}$, $\igp, \jgp, \iep, \jep$, prompting function $f_{\text{prompt}}$
\\
\textbf{Input:} test example $\vx$\\
%  \textbf{Initialization:} initialize $\theta$ by training on $\mathcal{D}_l$\; initialize $U$ by spectral embedding of $f_\theta(X)$\\
    Generate query feature $q(\vx)$\\
    Matching for the index of \eprompt via $t_{\vx} = \operatorname{argmin}_t\  \gamma(q(\vx), \vk_t)$\\
    Select the task-specific \eprompt $\ve_{t_{\vx}}$\\
    Generate the prompted architecture $f_{\vg, \ve_{t_{\vx}}}$: attach $\vg$ and $\ve_{t_{\vx}}$ to $\igp$-th to $\jgp$-th and $\iep$-th to $\jep$-th MSA layers respectively, with $f_{\text{prompt}}$. \\
    Prediction: $f_{\vg, \ve_{t_{\vx}}}(\vx)$

 \caption{\method at test time}
 \label{alg:dualprompt_test}
\end{algorithm}

\section{Algorithms for \method} \label{app:algorithm}
The training and test time Algorithms for \method are illustrated in Algorithm~\ref{alg:dualprompt_train} and~\ref{alg:dualprompt_test}, respectively.

\section{Experimental details} \label{app:exp_details}
%\zizhaoz{Likely goto supp.}\sayna{yeah definitely}
For our method, \method, we use a constant learning rate of $0.005$ usng Adam~\cite{kingma2014adam} optimizer with $\beta_1= 0.9$ and $\beta_2=0.999$, and a batch size of $128$ for all benchmarks. For methods with the ViT architecture, all input images are resized to $224\times224$ and normalized to $[0, 1]$. Otherwise, we follow their original implementation in their paper. We set the balancing factor $\lambda = 1$ in~\eqref{eq:full_loss}. We train Split CIFAR-100 and 5-datasets for $5$ epochs per task and Split ImageNet-R for $50$ epochs to ensure models converge properly for each task, thus the issue of forgetting is disentangled from possible underfitting~\cite{buzzega2020dark}. We further sample $20\%$ of the training set of Split ImageNet-R as a validation set for searching the optimal $\igp, \jgp$, $\iep, \jep$, and empirically set $\igp=1, \jgp=2$, $\iep=3, \jep=5$ for all the setting, since we discover they perform consistently well for all datasets. Following the suggestion of prompt length by \cite{wang2021learning}, we set $L_g=5$ and $L_e=20$ for all datasets as well, and we further verify the correctness of this choice in Appendix~\ref{app:prompt_length}. Note that for fair comparison, we set $M=30, L_p=20, N=5$ for L2P, which leads to similar amount of parameters as \method. %\hl{ For the prompting funciton $\pfunc$, we use Pro-T for Split CIFAR-100 and Split ImageNet-R, and Pre-T for 5-datasets, we will show the influence of $\pfunc$ in }. \zizhaoz{If you have a seperate section for 5-dataset to discuss this, I suggest to put to that place}

To ensure fair comparison, every aforementioned methods start from the same ImageNet pre-trained ViT-B/16 \cite{vit}, following the setting in~\cite{wang2021learning}. We carefully re-implement these method and use hyper-parameters by referring to their original source code. Moreover, we make the pre-trained model fully trainable for all methods (except L2P and \method), as we empirically observe they could not learn as good with a frozen backbone due to limited learning capacity. 

\section{Evaluation metrics} \label{app:eval_metrics}
Let $S_{t, \tau}$ be teh evaluation score, \eg, classification accuracy on the $\tau$-th task  after training on the $t$-th task. After the model finishes training on the $t$-th task, we compute the Average Accuracy ($A_t$) and Forgetting ($F_t$) as follows:
\begin{equation*}
\begin{aligned}
&A_{t}=\frac{1}{t} \sum_{\tau=1}^{t} S_{t, \tau}\\
&F_{t}=\frac{1}{t-1} \sum_{\tau=1}^{t-1} \max _{\tau^{\prime} \in\{1, \cdots, t-1\}}\left(S_{\tau^{\prime}, \tau}-S_{t, \tau}\right)
\end{aligned}
\end{equation*}
Note that Average Accuracy is the overall evaluation metric for continual learning, which includes two aspects: greater learning capacity and less catastrophic forgetting, while Forgetting only serves as a measure of catastrophic forgetting.

\section{Details of comparing methods} \label{app:comparing_methods}
\begin{itemize}
    \item \textbf{Regularization-based methods.} \textsf{EWC} \cite{kirkpatrick2017overcoming} and \textsf{LwF} \cite{li2017learning} are representative regularization-based methods that are widely compared.
    \item \textbf{Rehearsal-based methods.} \textsf{ER} \cite{chaudhry2019tiny,hayes2019memory}, \textsf{GDumb} \cite{prabhu2020gdumb}, \textsf{BiC} \cite{wu2019large}, \textsf{DER++} \cite{buzzega2020dark} and \textsf{Co$^2$L} \cite{cha2021co2l}. As earlier methods, ER and GDumb achieve very strong performance not only in their own work, but in later literature \cite{mai2021online,buzzega2020dark} as well. BiC is also a strong method in the class-incremental learning setting. DER++ and Co$^2$L are the latest SOTA methods. We chose a medium and a large buffer size for these rehearsal-based methods, based on recommendation of prior work~\cite{buzzega2020dark,mehtaempirical,cha2021co2l,wang2021learning}.
    \item \textbf{Prompt-based method.} \textsf{L2P}~\cite{wang2021learning} is the current state-of-the-art prompt-based method, we configure \method to have the similar amount of additional parameters as L2P for fair comparison.
    \item \textbf{Architecture-based methods.} All these methods are based on ResNet-18, as recommended in the original work. We either directly taken reported results in their original work or strictly follow the original implementation and hyper-parameter settings of these methods to reproduce the results. Some other well-known methods \cite{ke2020continual,ebrahimi2020adversarial,pham2020contextual} are not compared here because they either have been outperformed by our compared methods or only verified on simpler task-incremental setting.
\end{itemize}

\begin{figure}[t]
\centering 
\includegraphics[width=0.8\linewidth]{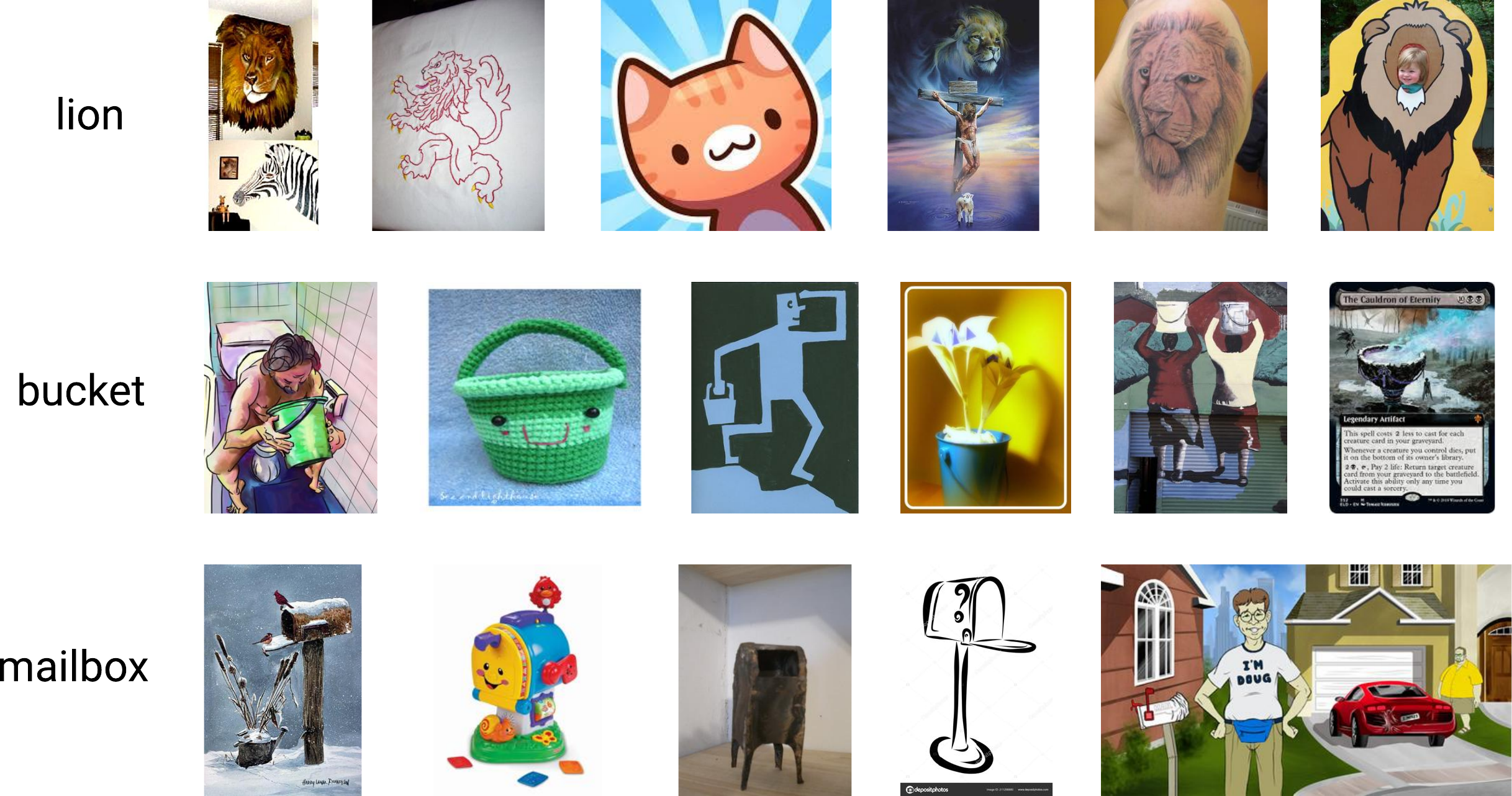} 
\vspace{-.3cm}
\caption{Representative examples from Split ImageNet-R.}
\label{fig:split_imr}
\end{figure}

\section{Split ImageNet-R: large intra-class diveristy} \label{app:split_imr}
Figure~\ref{fig:split_imr} shows some representative examples of three different classes from Split ImageNet-R. We can observe that although the images of the same row share the same label, they actually differ a lot. This observation accords well with the result in Figure~\ref{fig:intro} and Table~\ref{table:cifar_imagenet} that rehearsal-based methods require a large buffer size to perform well due to large intra-class diversity in Split ImageNet-R.

\section{Searching for multi-layered prompts} \label{app:multi_layer}
Based on the strategy in~\ref{sec:combiner}, we first search for the multi-layered \eprompt on the validation set of Split ImageNet-R. Due to the large search space, we made a simplified assumption that the MSA layers to attach prompts should be contiguous. Moreover, since $\iep = \jep = 5$ is the best option in the single layer case, it is natural to include the 5-th layer when searching for multi-layered \eprompt. Nevertheless, we also include several cases when the $5$-th layer is not included for completeness.

\begin{table}[t]
    \begin{minipage}{.48\textwidth}
        \centering 
        \captionof{table}{Searching results for multi-layered \eprompt.}
        \vspace{0.2cm}
            \scalebox{0.95}{
            \begin{tabular}[t]{cc|c}
            \toprule 
             $i_g$ &  $j_g$ & Avg. Acc\\
            %  &
            \midrule
            5 & 5 & 65.55 \\
            3 & 4 & 65.76 \\
            4 & 5 & 66.59 \\
            5 & 6 & 66.12 \\
            \bf3 & \bf5 & \bf67.12 \\
            4 & 6 & 66.41 \\
            5 & 7 & 64.53 \\
            1 & 12 & 67.09 \\
            \bottomrule
            \end{tabular}
            }
            \label{table:multi-eprompt}
    \end{minipage}
    \hspace{0.2cm}
    \begin{minipage}{.48\textwidth}
        \centering 
        \vspace{-0.7cm}
        \captionof{table}{Searching results for multi-layered \gprompt.}
        \vspace{0.2cm}
        \scalebox{0.95}{
            \begin{tabular}[t]{cc|c}
            \toprule 
             $i_e$ &  $j_e$ & Avg. Acc\\
            %  &
            \midrule
            2 & 2 & 67.70 \\
            \bf 1 & \bf 2 & \bf 68.46 \\
            1 & 3 & 67.79 \\
            1 & 5 & 67.73 \\
            6 & 8 & 65.10 \\
            1 & 12 & 63.13 \\
            \bottomrule
            \end{tabular}
        }
        \label{table:multi-gprompt}
    \end{minipage}
\end{table}

% \begin{table*}[h]
% \small
% \caption{Searching results for multi-layered \eprompt.}
% \label{table:multi-eprompt}
% \begin{center}
% \begin{tabular}{cc|c}
% \toprule 
%  $i_g$ &  $j_g$ & Avg. Acc\\
% %  &
% \midrule
% 5 & 5 & 65.55 \\
% 3 & 4 & 65.76 \\
% 4 & 5 & 66.59 \\
% 5 & 6 & 66.12 \\
% \bf3 & \bf5 & \bf67.12 \\
% 4 & 6 & 66.41 \\
% 5 & 7 & 64.53 \\
% 1 & 12 & 67.09 \\
% \bottomrule
% \end{tabular}
% \end{center}
% \end{table*}
We discover that $\iep = 3, \jep = 5$ yields the best performance in terms of average accuracy. Note that when we attach \eprompt to every MSA layer ($\iep = 1, \jep = 12$), it actually leads to comparable accuracy. However, we still choose $\iep = 3, \jep = 5$ since it has less additional parameters.

We then fix $\iep = 3, \jep = 5$, and search for multi-layered \gprompt, given that we have $\igp = \jgp = 2$ yields the best performance as a single-layered \eprompt. We conduct similar searching process, with a preference of including the 2nd layer. The searching results are shown in Table~\ref{table:multi-gprompt}.

% \begin{table*}[h]
% \small
% \caption{Searching results for multi-layered \gprompt.}
% \label{table:multi-gprompt}
% \begin{center}
% \begin{tabular}{cc|c}
% \toprule 
%  $i_e$ &  $j_e$ & Avg. Acc\\
% %  &
% \midrule
% 2 & 2 & 67.70 \\
% \bf 1 & \bf 2 & \bf 68.46 \\
% 1 & 3 & 67.79 \\
% 1 & 5 & 67.73 \\
% 6 & 8 & 65.10 \\
% 1 & 12 & 63.13 \\
% \bottomrule
% \end{tabular}
% \end{center}
% \end{table*}
We discover that $\igp = 1, \jgp = 2$ leads to best average accuracy. Moreover, simply share all MSA layers results in overall negative effect on the accuracy.

Although our search strategy is not exhaustive, we find the combination of $\igp = 1, \jgp = 2, \iep = 3, \jep = 5$ works quite well for all benchmark datasets.

\section{Searching for prompt length} \label{app:prompt_length}
%%%%%%%%%%%%%%%%%%%%%%%%%%%%%%%%%%%%%%%%%%%%%%%%%%%%%%%%%%%%%%%%%%%%%%%%%%%%%%%
%%%%%%%%%%%%%%%%%%%%%%%%%%%%%%%%%%%%%%%%%%%%%%%%%%%%%%%%%%%%%%%%%%%%%%%%%%%%%%%
Following the suggestion by \cite{wang2021learning}, we use set the base length of prompts as $5$, and perform grid search on the lengths of \gprompt and \eprompt from $\{5, 10, 20, 40\} \times \{5, 10, 20, 40\}$, based on the optimal positions obtained in the previous step.

\begin{figure}[t]
\centering 
\includegraphics[width=0.35\linewidth]{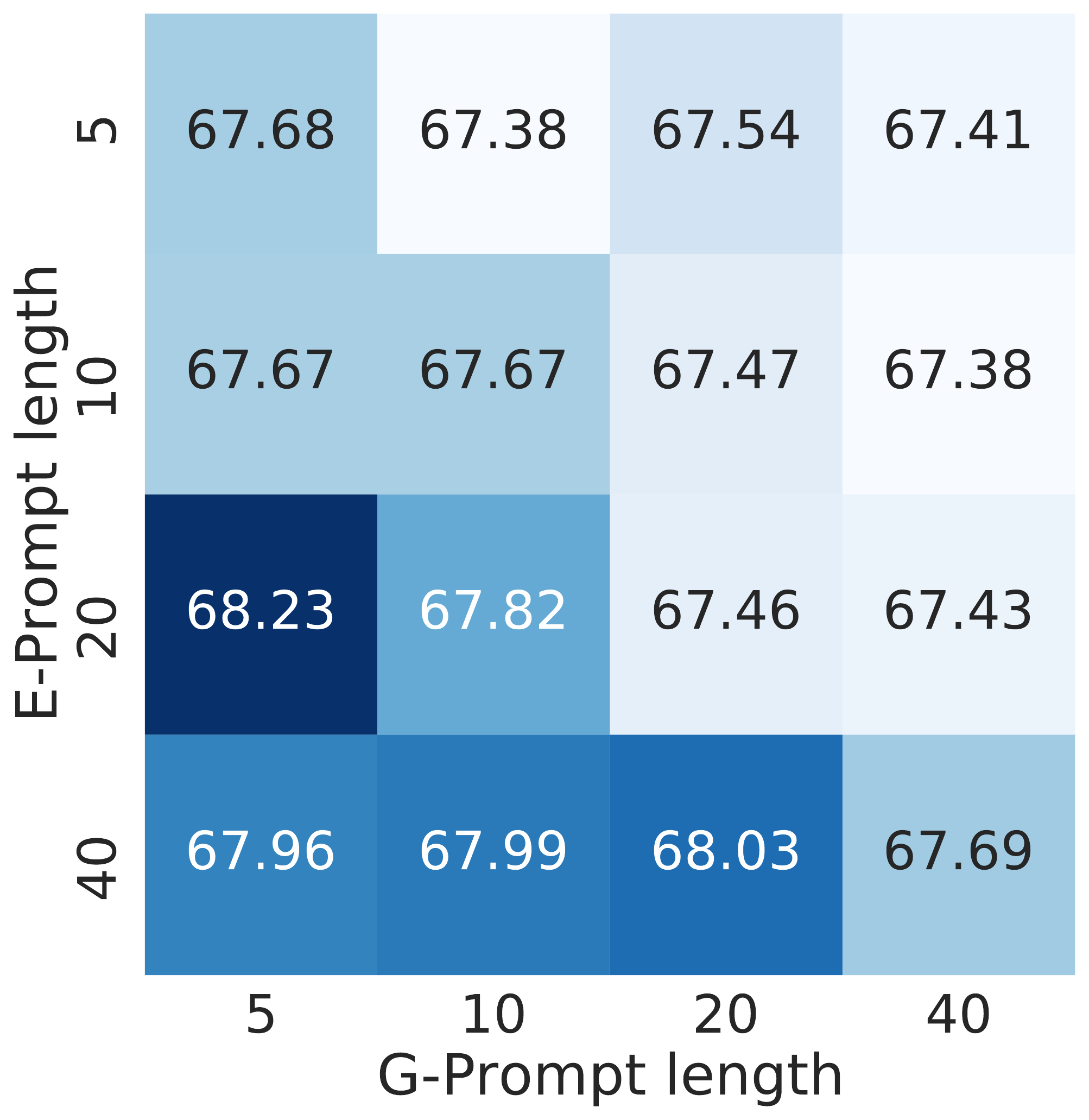} 
\vspace{-.3cm}
\caption{Grid search on prompt length.}
\label{fig:prompt_length}
\end{figure} 

As shown in Figure~\ref{fig:prompt_length}, we indeed verify $L_g = 5, L_e=20$ is the optimal choice on the validation set of Split ImageNet-R. We also empirically observe this configuration works well on other datasets, thus we use this combination of prompt lengths for other datasets in experiments showed in the main text.

\begin{table}[t]
\small
\caption{Additional results on 5-datasets \cite{ebrahimi2020adversarial}.}
\label{table:5datasets}
\begin{center}
\scalebox{0.8}{
\begin{tabular}{l||c|cc}
\toprule 
 \multirow{2}{*}{\textbf{Method}} & \multirow{2}{*}{\textbf{Buffer size}} & \multicolumn{2}{c}{\textbf{5-datasets}} \\
%  &
& &  Avg. Acc ($\uparrow$) & Forgetting ($\downarrow$) \\
\midrule
  ER  & \multirow{6}{*}{250} & {80.32\scriptsize{$\pm$0.55}} & 15.69\scriptsize{$\pm$0.89} \\
  BiC  && 78.74\scriptsize{$\pm$1.41} & 21.15\scriptsize{$\pm$1.00} \\
  % GDumb  && 56.99\scriptsize{$\pm$0.06} & - \\
 DER++  && 80.81\scriptsize{$\pm$0.07} & 14.38\scriptsize{$\pm$0.35} \\
 Co$^2$L   && 82.25\scriptsize{$\pm$1.17} & 17.52\scriptsize{$\pm$1.35} \\
 % L2P-R \cite{wang2021learning} & & {84.21\scriptsize{$\pm$0.53}} & {7.72\scriptsize{$\pm$0.77}} && {85.56\scriptsize{$\pm$0.95}} & {4.22\scriptsize{$\pm$0.03}} \\
 \midrule
 ER  &\multirow{6}{*}{500}& {84.26\scriptsize{$\pm$0.84}} & 12.85\scriptsize{$\pm$0.62} \\
 BiC  && 85.53\scriptsize{$\pm$2.06} & 10.27\scriptsize{$\pm$1.32} \\
 % GDumb  && 70.76\scriptsize{$\pm$0.12} & - \\
 DER++  && 84.88\scriptsize{$\pm$0.57} & 10.46\scriptsize{$\pm$1.02} \\
 Co$^2$L  && 86.05\scriptsize{$\pm$1.03} & 12.28\scriptsize{$\pm$1.44} \\
 \midrule
 % FT-seq-frozen & \multirow{6}{*}{0} & 17.72\scriptsize{$\pm$0.34} & 59.09\scriptsize{$\pm$0.25} & \multirow{6}{*}{0} & 39.49\scriptsize{$\pm$0.12} & 42.62\scriptsize{$\pm$0.20} \\ 
 FT-seq & \multirow{5}{*}{0} & 20.12\scriptsize{$\pm$0.42} & 94.63\scriptsize{$\pm$0.68} \\
 EWC  && 50.93\scriptsize{$\pm$0.09} & 34.94\scriptsize{$\pm$0.07} \\
 LwF  && 47.91\scriptsize{$\pm$0.33} & 38.01\scriptsize{$\pm$0.28} \\
 {L2P}  &&81.14\scriptsize{$\pm$0.93} & {4.64\scriptsize{$\pm$0.52}} \\
 \bf{\method} && \bf 88.08\scriptsize{$\pm$0.36} & \bf 2.21\scriptsize{$\pm$0.69} \\
\midrule
Upper-bound & - & 93.93\scriptsize{$\pm$0.18} & - \\
\bottomrule
\end{tabular}
}
\end{center}
\end{table}

\section{Additional results on 5-datasets} \label{app:5-datasets}
% \zifeng{maybe put this part to supplement? } \sayna{yeah definitely. Mention you have it in the main paper though}
For completeness, we also demonstrate the effectiveness of our method on 5-datasets~\cite{ebrahimi2020adversarial}, which is a collection of five diverse image classification datasets, CIFAR-10~\cite{krizhevsky2009learning}, MNIST~\cite{lecun1998mnist}, Fashion-MNIST~\cite{xiao2017fashion}, SVHN~\cite{netzer2011reading}, and notMNIST~\cite{notmnist}. Despite the simplicity of each task in 5-datasets, the benchmark mimics the real-world setting where task diversity is large, thus contributing to a more comprehensive evaluation of CL methods. Since each task in 5-dataset is relatively easier than that of Split CIFAR-100 and ImageNet-R, rehearsal-based methods generally require smaller buffer size to perform well. However, a buffer size of 500 is already considered large for 5-datasets~\cite{ebrahimi2020adversarial,mehtaempirical}. Although \method still consistently outperforms competing methods, we argue that in real world continual learning scenarios, it is rare that tasks are too diverse: for example, we don't expect a digit classifier to continually learn to classify animals. Thus, we only show the performance on 5-datasets as a proof-of-concept that our method works well when task diversity is large.

\begin{table}[t]
    \centering
    \caption{Comparison between \method with query strategy introduced in Section~\ref{sec:g-e-prompt}, and with the perfect match (known test time task identity) on Split ImageNet-R.}
    \scalebox{0.8}{
    \begin{tabular}{l|c|c|c}
    \toprule 
     \textbf{Method} & \bf Matching Acc & \textbf{Avg Acc ($\uparrow$)} & \textbf{Forgetting ($\downarrow$)}  \\
    \midrule
    \method-Query & 55.8 & 68.13 & 4.68 \\
    \method-Perfect Match & 100 & 71.97 & 3.95 \\
    \bottomrule
    \end{tabular}
    }
    \label{table:perfect_match}
\end{table}

\section{Relationship between query accuracy and performance} \label{app:perfect_match}
To demonstrate the relationship between query accuracy and performance, we compare \method with the query strategy introduced in ~\ref{sec:g-e-prompt} to \method with known test time task identity to select \eprompt. The result is shown in Table~\ref{table:perfect_match}. 
Interestingly, although the matching accuracy is not that great, \method is quite robust to it and still achieves an accuracy very close to perfect match. We contribute this robustness to mismatching to the design of our method. First, the task-invariant instruction captured in \gprompt still remains useful for prediction even if \eprompt is noisy. Second, the query strategy is based on the input feature, thus implicitly taking into account task similarity. Even when mismatching happens, our method tends to choose the \eprompt from one of the most similar tasks. We also note that their is still forgetting even if we use ground truth task identity to select the corresponding task-specific \eprompt. This part of forgetting results from the bias in the final softmax classification head~\cite{mai2021online,prabhu2020gdumb,zeno2018task}, a common issue in class-incremental learning that could be mitigated in parallel.

\end{document}

%% file: math_commands.tex
\usepackage{amsmath,amsfonts,bm}

\def\eqref#1{equation~\ref{#1}}
\def\1{\bm{1}}

\newcommand{\test}{\mathcal{D_{\mathrm{test}}}}

\def\ve{{\bm{e}}}

\def\vg{{\bm{g}}}
\def\vh{{\bm{h}}}

\def\vk{{\bm{k}}}

\def\vp{{\bm{p}}}

\def\vx{{\bm{x}}}

\DeclareMathAlphabet{\mathsfit}{\encodingdefault}{\sfdefault}{m}{sl}
\SetMathAlphabet{\mathsfit}{bold}{\encodingdefault}{\sfdefault}{bx}{n}

\def\sR{{\mathbb{R}}}